\documentclass[lettersize,journal]{IEEEtran}
\usepackage{amsmath,amssymb,amsfonts}
\usepackage{algorithm}
\usepackage{array}
\usepackage[caption=false,font=footnotesize,labelfont=rm,textfont=rm]{subfig}
\usepackage{textcomp}
\usepackage{makecell}
\usepackage{url}
\usepackage{verbatim}
\usepackage{graphicx}
\usepackage{cite}
\usepackage{hyperref} 
\usepackage{bigstrut}
\usepackage{multicol}
\usepackage{multirow} 
\usepackage{siunitx}
\usepackage{pifont}
\usepackage{bbding}    
\usepackage{xcolor}
\usepackage{colortbl}
\newcommand{\best}[1]{\textbf{#1}}
\newcommand{\second}[1]{#1}
\usepackage{algpseudocode}
\usepackage{booktabs,tabularx}
\hypersetup{
    colorlinks=true, 
    linkcolor=blue, 
    citecolor=green, 
    urlcolor=blue 
}
\begin{document}

\title{

S$^3$AM: A Single-Stream SAM with Reliability-Calibrated Frequency Adapter for Multi-modal Salient Object Detection
}

\author{Ruichao Hou, Boyue Xu, Tongwei Ren, Dongming Zhou, Gangshan Wu, and Jinde Cao, \IEEEmembership{Fellow, IEEE}

\thanks{This work was supported by the National Natural Science Foundation of China (92582103, 62576098), the Fundamental and Interdisciplinary Disciplines Breakthrough Plan of the Ministry of Education of China (No. JYB2025XDXM118), the ``111 Center'' (No. B26023), and the Collaborative Innovation Center of Novel Software Technology and Industrialization.
\emph{(Corresponding authors: Tongwei Ren)} }
\thanks{Ruichao Hou is with the School of Elite Biomedical Engineers and the Institute for Interdisciplinary Intelligent Pharmacy, China Pharmaceutical University, Nanjing 211198, China (e-mail: rchou@cpu.edu.cn).}
\thanks{Boyue Xu, Tongwei Ren, and Gangshan Wu are with the State Key Laboratory for Novel Software Technology, Nanjing University, Nanjing 210008, China (e-mail: xuby@smail.nju.edu.cn; rentw@nju.edu.cn; gswu@nju.edu.cn).}
\thanks{Dongming Zhou is with the School of Information Science and Engineering, Yunnan University, Kunming 650091, China (e-mail: zhoudm@ynu.edu.cn).}
\thanks{Jinde Cao is with the School of Mathematics, Southeast University, Nanjing 211189, China, and also with the Purple Mountain Laboratories, Nanjing 211111, China (email: jdcao@seu.edu.cn).}

\thanks{Manuscript received **** **, 2026; revised **** **, 2026.}
}

\markboth{Submitted to IEEE Journals/Transactions}%
{Hou \MakeLowercase{\textit{et al.}}: S\textsuperscript{3}AM for Multi-modal Salient Object Detection}


\maketitle

\begin{abstract}
Vision foundation models have recently advanced multi-modal salient object detection (MSOD) through parameter-efficient tuning and prompt learning. However, existing Segment Anything Model (SAM)-adapted MSOD methods often rely on dual-stream encoders or auxiliary prompt generators, leading to redundant computation. Although a single-stream alternative can reduce this cost, early fusion may also propagate noisy or misaligned auxiliary high-frequency cues through the backbone.
In this paper, we propose a novel single-stream framework that integrates reliability-calibrated frequency adaptation into the adopted SAM backbone for MSOD. It avoids duplicated foundation backbones while explicitly controlling auxiliary frequency injection.
Specifically, we design a mixture of frequency experts module, which uses the stationary wavelet transform to decompose each modality and aggregate cross-modal frequency information. We further introduce a reliability-calibrated frequency adapter with a dual-gate calibration mechanism, which selectively propagates the calibrated residual across transformer stages while jointly controlling its injection strength and cross-modal reliability. A hypernetwork-guided semantic-structural decoder then combines semantic mask features from the adopted backbone with Mamba-based structural detail recovery.
Comprehensive experiments on RGB-D, RGB-T, and RGB-NIR salient object detection benchmarks validate that the proposed framework achieves competitive performance with only 12.20M trainable parameters, accounting for 5.4\% of the total parameters. The code will be available at \url{https://github.com/xuboyue1999/SSSAM}.

\end{abstract}


\begin{figure}[t]
\centering
  \includegraphics[width=0.48\textwidth]{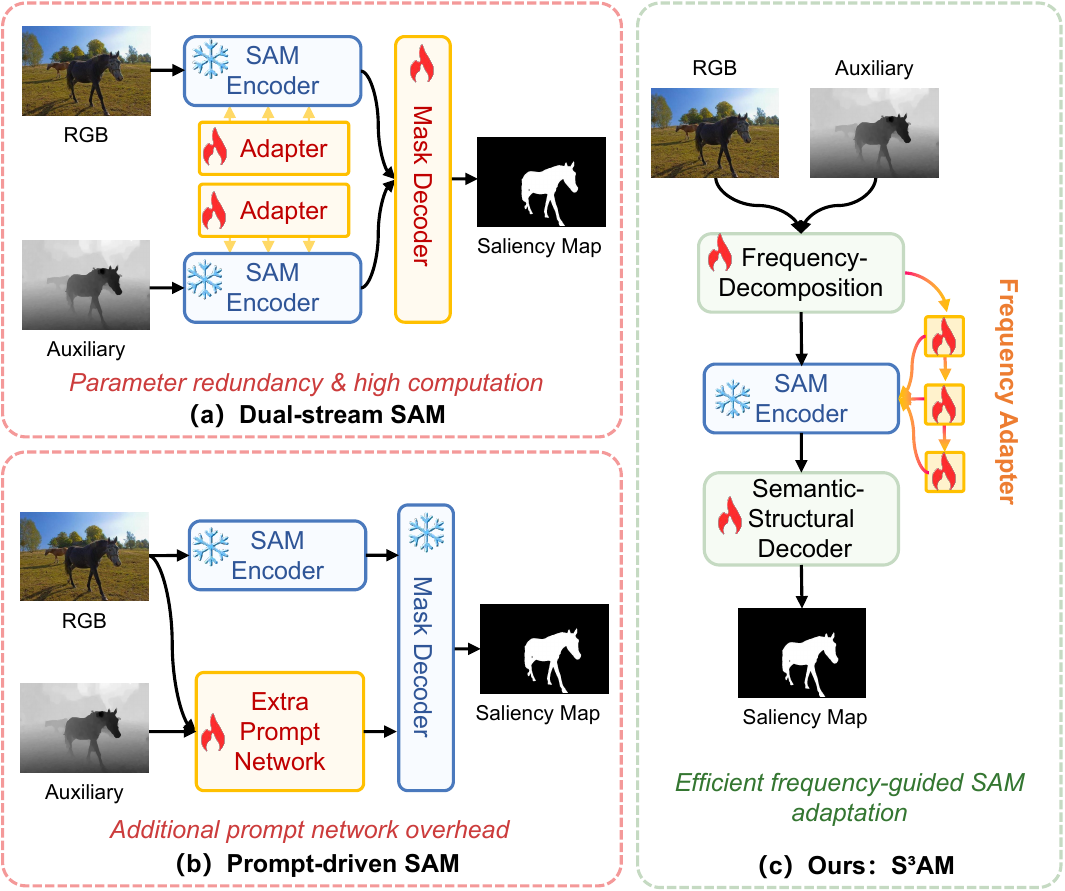}   
  \caption{Comparison of SAM adaptation paradigms for multi-modal salient object detection. (a) Dual-stream SAM encodes RGB and auxiliary inputs with two separate SAM encoders, resulting in redundant foundation-backbone computation. (b) Prompt-driven SAM converts auxiliary cues into extra prompts through an additional prompt network. (c) Our S$^3$AM performs early frequency-aware fusion and adapts a single frozen SAM backbone with reliability-calibrated frequency adapters.} 
  \label{fig1}  
\end{figure}

\IEEEPARstart{M}{ulti-modal} salient object detection (MSOD) aims to identify and highlight the most visually distinctive regions in an image by leveraging complementary information from multiple sensing modalities. By integrating inputs such as RGB, depth (RGB-D), thermal (RGB-T), or near-infrared (RGB-NIR) data, MSOD enhances feature representation under challenging conditions like low illumination, background clutter, or visual camouflage. This enhanced robustness makes MSOD a crucial component in various downstream tasks, including object tracking and video object detection~\cite{wang2022mfgnet,wang2026multi}, virtual reality rendering~\cite{gao2025saliency}, and image retrieval~\cite{lu2025image}. The additional structural or semantic cues offered by multiple modalities enable more reliable saliency prediction in real-world scenarios.

Most existing MSOD methods adopt dual-stream architectures, where two modality-specific backbones independently extract features that are later fused through specialized fusion modules. The goal of such designs is to exploit both the consistency and complementarity of different modalities to construct robust multi-modal representations. However, dual-stream processing inevitably duplicates a large portion of the visual backbone, and this redundancy becomes more pronounced when vision foundation models are introduced. Meanwhile, the limited size of publicly available multi-modal datasets, constrained by the cost of aligned data acquisition and labor-intensive annotation, poses a major challenge for training specialized architectures from scratch.
To mitigate the data limitation, recent works have begun incorporating SAM~\cite{ravi2024sam} into the MSOD task, leveraging its strong zero-shot generalization under limited supervision. Two representative adaptation paradigms have emerged, as shown in Fig.~\ref{fig1}. The first is parameter-efficient fine-tuning, where lightweight adapters, LoRA modules, or task-specific adaptation modules are inserted into the frozen encoder to inject task-specific knowledge~\cite{chen2023sam,kansam,liu2026samsod}. The second is prompt learning, in which an auxiliary SOD network or prompt generator produces saliency-aware prompts that guide the unmodified SAM toward task-relevant segmentation results~\cite{hypsam}. Despite their effectiveness, these paradigms often depend on duplicated modality-specific processing, additional prompt generation, or task-specific optimization, introducing computational overhead that limits deployment in real-time or resource-constrained environments.

An appealing alternative is to adapt SAM in a single-stream manner, where RGB and auxiliary modalities are fused early and processed by a unified encoder. This design directly reduces architectural redundancy, but it also changes the reliability requirement of multi-modal fusion: once auxiliary cues are injected at an early stage, their errors can propagate through the entire foundation backbone. This issue is particularly evident for high-frequency information. Although high-frequency cues are crucial for preserving object boundaries, thin structures, and small salient regions, they are not uniformly trustworthy across modalities. Depth maps may contain holes or shifted boundaries, thermal images may be blurred, and NIR responses may emphasize modality-specific textures. Blindly injecting such high-frequency cues can therefore contaminate early features and sharpen wrong contours rather than improve saliency localization. Therefore, the key challenge is not only how to build an efficient single-stream SAM framework, but also how to make early frequency injection reliable.

To this end, we propose S$^3$AM, a novel and efficient single-stream SAM adaptation framework that introduces reliability-calibrated frequency adaptation for multi-modal SOD tasks. Unlike previous dual-stream or prompt-driven approaches, S$^3$AM processes multi-modal inputs within a unified encoder to avoid redundant foundation backbones, and further controls the propagation of auxiliary high-frequency cues through reliability-calibrated frequency adaptation.
Accordingly, S$^3$AM is organized around three operations. First, the Mixture of Frequency Experts (MoFE) constructs image-conditioned frequency cues for early single-stream fusion. Second, the Reliability-Calibrated Frequency Adapter (RCFA) evaluates and calibrates them during stage-wise propagation, suppressing unreliable residuals through dual-gate control of contextual injection and RGB--auxiliary reliability. Third, the lightweight Hypernetwork-guided Semantic-Structural Decoder (HSSD) translates the calibrated features into saliency masks by coupling the adopted SAM backbone's hypernetwork-based semantic prior with structural detail recovery under joint saliency-edge supervision.
Extensive experiments on multiple MSOD benchmarks demonstrate that S$^3$AM achieves competitive performance while maintaining parameter-efficient adaptation.

In summary, our main contributions are as follows:

\begin{itemize}
\item We propose a parameter-efficient single-stream SAM adaptation framework for multi-modal salient object detection that avoids duplicated foundation backbones while controlling noisy auxiliary frequency injection, thereby reducing architectural redundancy without sacrificing cross-modal interaction.
\item We propose a Mixture of Frequency Experts to construct image-conditioned frequency cues for early fusion, thereby supplying diverse structural information to the unified encoder.
\item We propose a Reliability-Calibrated Frequency Adapter that uses dual-gate calibration to selectively propagate useful frequency residuals and suppress unreliable ones across transformer stages, thereby preventing noisy auxiliary details from contaminating subsequent features.
\item We employ a Hypernetwork-guided Semantic-Structural Decoder that combines the adopted SAM backbone's semantic mask prior with Mamba-based detail recovery under joint mask and edge supervision, thereby improving semantic completeness and boundary accuracy.

\end{itemize}

\section{Related Work}
\subsection{Multi-modal Salient Object Detection}
MSOD enhances RGB-based SOD by integrating complementary cues from modalities such as depth, thermal, or near-infrared (NIR), which provide richer spatial or semantic context for robust object perception.
Recent MSOD methods are generally categorized into single-stream, dual-stream, and triple-stream architectures. Single-stream designs, \emph{e.g.}, OSRNet~\cite{huo2022real}, emphasize efficiency by performing early fusion through simple operations like concatenation and element-wise addition.

Dual-stream frameworks remain the most widely adopted due to their effectiveness in extracting and aligning modality-specific features. For instance, BTNet~\cite{ren2025bio} leverages a biologically inspired two-branch design to process RGB and depth in parallel. DiMSOD~\cite{zhang2025dimsod} formulates multi-modal salient object detection as a conditional mask generation problem and introduces a diffusion-based framework to progressively refine saliency predictions.
Triple-stream models, such as MDBIFNet~\cite{xie2023cross}, introduce an interaction stream to further refine cross-modal feature integration. 
Despite the performance gains brought by various fusion strategies, the effectiveness of existing MSOD models remains constrained by the limited scale and diversity of available multi-modal datasets. Consequently, incorporating vision foundation models to boost generalization under limited supervision has become an increasingly promising direction.

At the same time, single-stream architectures remain appealing for real-time or resource-constrained applications due to their lightweight design and lower computational cost. Motivated by these insights, we investigate how to combine a single-stream architecture with SAM while avoiding duplicated foundation-backbone computation.

\begin{figure*}[t]
  \centering
  \includegraphics[width=\textwidth]{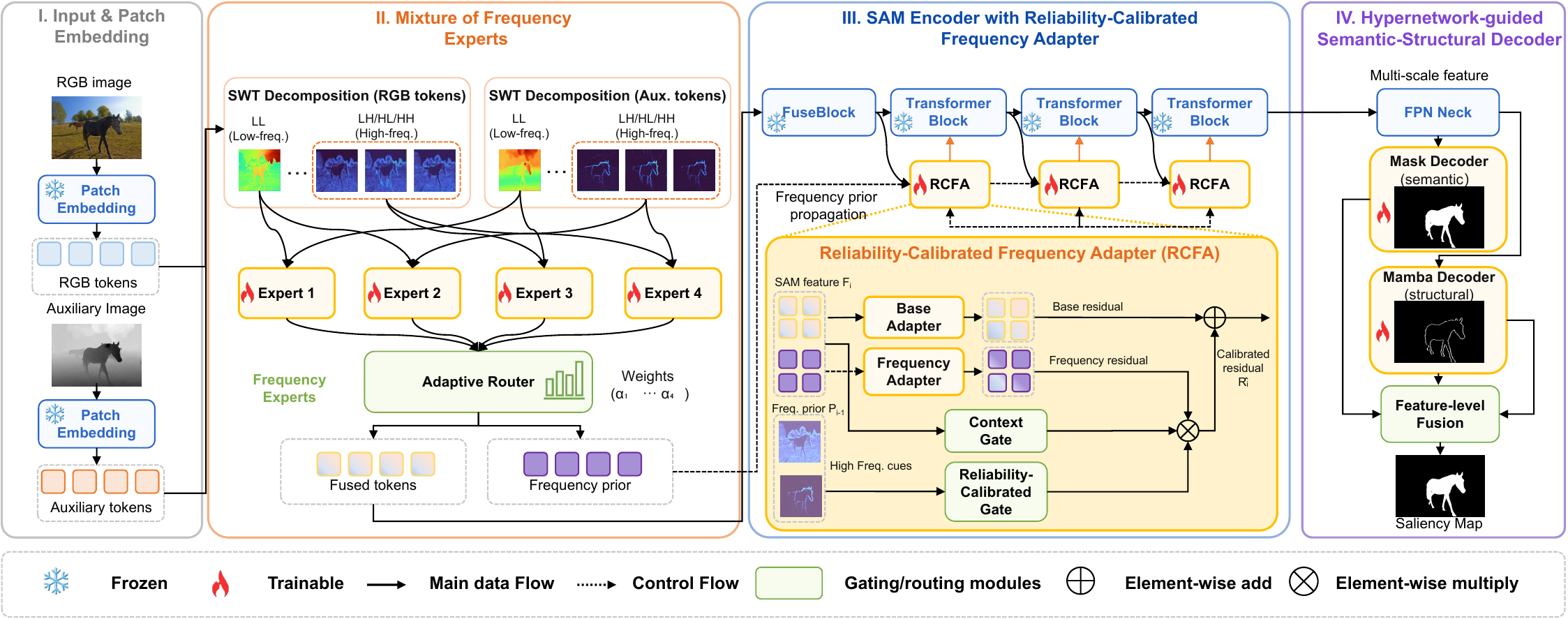}   
  \caption{Framework of the proposed S$^3$AM. Given RGB and auxiliary features $\mathcal{F}^{\mathrm{rgb}}$ and $\mathcal{F}^{\mathrm{aux}}$, MoFE performs SWT-based frequency decomposition and cross-modal expert aggregation to obtain $\mathcal{F}^{\mathrm{fused}}$, which is added to $\mathcal{F}^{\mathrm{rgb}}$ before the first block of the adopted SAM backbone, and the initial candidate frequency prior $\mathcal{P}_0$. At RCFA stage $i$, the current encoder feature $\mathcal{F}_i$ and incoming frequency prior $\mathcal{P}_{i-1}$ produce a calibrated residual $\mathcal{F}^{\mathrm{res}}_i$, which is injected to form $\mathcal{F}^{\mathrm{out}}_i$ and propagated as the calibrated frequency prior $\mathcal{P}_i$. The resulting multi-scale encoder features are decoded by HSSD, which combines the adopted SAM backbone's hypernetwork-guided semantic prediction with Mamba-based structural detail recovery.}   
  \label{fig:framework}  
\end{figure*}

\subsection{SAM and Its Application}
The adopted SAM backbone~\cite{ravi2024sam} is a foundation model for image and video segmentation that provides strong general-purpose visual representations and supports prompt-based segmentation from inputs such as points and bounding boxes.

Owing to its versatility, SAM has been widely extended to various visual domains. For instance, the SAM-Adapter~\cite{chen2023sam} incorporates lightweight adapters into the frozen SAM encoder, injecting domain-specific knowledge or visual priors for specialized tasks. Similarly, MedSAM~\cite{ma2024segment} adapts SAM for medical image segmentation by fine-tuning on a curated large-scale medical dataset. In remote sensing, SAMRS~\cite{wang2023samrs} leverages SAM to construct a large-scale satellite image segmentation dataset and shows that SAM can generalize to overhead imagery with appropriate prompt strategies.

Moreover, SAM has been adapted for multi-modal vision tasks. SAGE~\cite{wu2025every} introduces a multi-modality image fusion framework that distills semantic priors from SAM to improve visual quality and downstream task adaptability without requiring SAM during inference. For RGB-T SOD, SACNet~\cite{SACNET} studies alignment-free cross-modal interaction through semantics-guided asymmetric correlation. HyPSAM~\cite{hypsam} employs a dynamic fusion network to generate an initial saliency map and uses hybrid text, mask, and box prompts to guide SAM refinement. SAMSOD~\cite{liu2026samsod} further revisits SAM optimization for RGB-T salient object detection.
Zhai \emph{et al.}~\cite{zhai2025weakly} further propose a SAM-guided label refinement framework for weakly supervised RGB-T salient object detection. SSFam~\cite{10909610} introduces a framework built on SAM for scribble-supervised salient object detection across various modality combinations, leveraging modal-aware modulation and a siamese decoder to unify sparse-label learning and modality-adaptive segmentation. These works demonstrate SAM’s strong generalization ability and its flexibility to serve as a backbone or a semantic prior extractor across diverse and complex vision applications. However, existing MSOD methods often rely on dual-stream architectures or additional prompt generators, limiting computational efficiency and failing to fully exploit frequency-aware cues, which motivates our design of a streamlined, frequency-modulated MSOD framework based on SAM.

\section{Methodology}
\subsection{Overview}
We propose S$^3$AM, a reliability-calibrated frequency adaptation framework tailored for single-stream MSOD and built upon the adopted SAM backbone~\cite{ravi2024sam}. As shown in Fig.~\ref{fig:framework}, the design follows the causal motivation discussed above. To avoid duplicating foundation backbones, the unified encoder processes RGB and auxiliary modalities in a single stream. Since this requires early cross-modal fusion, MoFE constructs initial frequency cues before the transformer stages. Since these cues can contain unreliable local high-frequency responses, RCFA performs stage-wise calibration and propagation with dual-gate control over injection strength and RGB--auxiliary reliability. The lightweight Hypernetwork-guided Semantic-Structural Decoder then converts the calibrated features into saliency maps by preserving the adopted SAM backbone's hypernetwork-based semantic prior and recovering structural details. Given a paired input of RGB and an auxiliary modality (\emph{e.g.}, depth, thermal, or NIR), features are first extracted through a shared patch embedding layer of the adopted SAM backbone, decomposed and aggregated via MoFE, selectively refined through a sequence of RCFAs, and decoded by HSSD into structure-preserving saliency maps.

\begin{figure}[t]
  \centering
  \includegraphics[width=\linewidth]{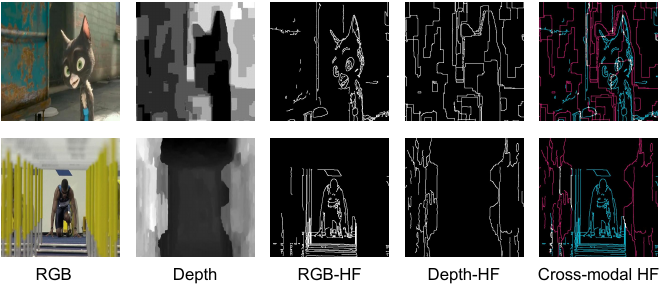}
  \caption{Motivation for reliability calibration. Raw RGB and depth inputs exhibit complementary yet locally inconsistent high-frequency responses. In the cross-modal relation map, cyan and magenta denote RGB-only and depth-only responses, respectively, and their coexistence reveals local disagreement.}
  \label{fig:hf-motivation}
\end{figure}

Fig.~\ref{fig:hf-motivation} provides an input-domain visualization of this motivation. High-frequency responses computed from raw RGB and depth images contain both cross-modally aligned boundaries and modality-specific or spatially misaligned structures. In the cross-modal relation map, cyan and magenta denote RGB-only and depth-only responses, respectively. Cyan--magenta coexistence indicates conflict.

\subsection{Mixture of Frequency Experts}

Conventional early fusion methods often inadequately leverage modality-specific cues and overlook fine-grained details. To address this, MoFE focuses on the first step of our frequency-aware design: extracting and aggregating cross-modal frequency cues before subsequent adapters calibrate them. It utilizes the Stationary Wavelet Transform (SWT)~\cite{demirel2010image} to perform multi-frequency decomposition and employs adaptive expert-driven aggregation to provide information-rich frequency cues.

Given RGB and auxiliary modality features  
$\mathcal{F}^{\mathrm{rgb}}, \mathcal{F}^{\mathrm{aux}} \in \mathbb{R}^{C \times H \times W}$,  
where the superscripts $\mathrm{rgb}$ and $\mathrm{aux}$ indicate the RGB and auxiliary modalities, and $C$, $H$, and $W$ denote the channel number, height, and width of the token feature map, respectively,
we apply SWT independently to both modalities to extract one low-frequency and three high-frequency subbands:

\begin{equation}
\begin{aligned}
\{ \mathcal{C}^{\mathrm{rgb}}_L, \mathcal{C}^{\mathrm{rgb}}_{LH}, \mathcal{C}^{\mathrm{rgb}}_{HL}, \mathcal{C}^{\mathrm{rgb}}_{HH} \} &= \mathrm{SWT}(\mathcal{F}^{\mathrm{rgb}}), \\
\{ \mathcal{C}^{\mathrm{aux}}_L, \mathcal{C}^{\mathrm{aux}}_{LH}, \mathcal{C}^{\mathrm{aux}}_{HL}, \mathcal{C}^{\mathrm{aux}}_{HH} \} &= \mathrm{SWT}(\mathcal{F}^{\mathrm{aux}}),
\end{aligned}
\end{equation}
where $\mathrm{SWT}(\cdot)$ denotes the stationary wavelet transform, $\mathcal{C}_L$ denotes the low-frequency subband, and $\mathcal{C}_{LH}$, $\mathcal{C}_{HL}$, $\mathcal{C}_{HH}$ correspond to vertical, horizontal, and diagonal high-frequency components, respectively.
For compact notation, we concatenate the three directional high-frequency subbands as:
\begin{equation}
\mathcal{C}^{m}_H =
\mathrm{Concat}(\mathcal{C}^{m}_{LH}, \mathcal{C}^{m}_{HL}, \mathcal{C}^{m}_{HH}),
\quad m \in \{\mathrm{rgb}, \mathrm{aux}\}.
\end{equation}
Here $m$ indexes the modality and $\mathrm{Concat}(\cdot)$ denotes channel-wise concatenation.

Subsequently, we construct four cross-modal frequency pairs by combining low- and high-frequency components across modalities:
\begin{equation}
\left\{
\begin{aligned}
\mathcal{C}_1 &= \mathrm{Concat}(\mathcal{C}^{\mathrm{rgb}}_L, \mathcal{C}^{\mathrm{aux}}_L), \\
\mathcal{C}_2 &= \mathrm{Concat}(\mathcal{C}^{\mathrm{rgb}}_L, \mathcal{C}^{\mathrm{aux}}_H), \\
\mathcal{C}_3 &= \mathrm{Concat}(\mathcal{C}^{\mathrm{rgb}}_H, \mathcal{C}^{\mathrm{aux}}_L), \\
\mathcal{C}_4 &= \mathrm{Concat}(\mathcal{C}^{\mathrm{rgb}}_H, \mathcal{C}^{\mathrm{aux}}_H),
\end{aligned}
\right.
\end{equation}
Here, $\mathcal{C}_1$ concatenates the low-frequency components of RGB and the auxiliary modality, providing a coarse cross-modal structural cue. $\mathcal{C}_2$ combines the RGB low-frequency component with the auxiliary high-frequency component, whereas $\mathcal{C}_3$ combines the RGB high-frequency component with the auxiliary low-frequency component. $\mathcal{C}_4$ concatenates the high-frequency components from both modalities and therefore captures their joint high-frequency cue. These four complementary frequency pairs are supplied to separate experts for adaptive enhancement.
Each frequency pair $\mathcal{C}_n$ is then passed through a dedicated expert module $E_n(\cdot)$ for enhancement:
\begin{equation}
\widehat{\mathcal{C}}_n = E_n(\mathcal{C}_n), \quad n \in \{1, 2, 3, 4\},
\end{equation}
where $n$ is the expert index, $E_n(\cdot)$ denotes the $n$-th frequency expert, and $\widehat{\mathcal{C}}_n$ is the corresponding enhanced frequency feature.

To adaptively aggregate these enhanced features, a lightweight router predicts normalized routing weights:
\begin{equation}
\boldsymbol{\alpha} =
\mathrm{Softmax}\left(\mathrm{Router}([\mathcal{C}_1,\mathcal{C}_2,\mathcal{C}_3,\mathcal{C}_4])\right),
\quad \sum_{n=1}^{4}\alpha_n = 1.
\end{equation}
Here $\boldsymbol{\alpha}=[\alpha_1,\alpha_2,\alpha_3,\alpha_4]$ denotes the routing-weight vector, $\alpha_n$ is the weight assigned to the $n$-th expert, $[\cdot]$ denotes feature concatenation for router input, and $\mathrm{Softmax}(\cdot)$ normalizes the expert weights.
The final fused representation is then computed as a weighted sum:

\begin{equation}
\mathcal{F}^{\mathrm{fused}} = \sum_{n=1}^{4} \alpha_n \widehat{\mathcal{C}}_n.
\end{equation}
The fused feature is added to the RGB feature before entering the first frozen transformer block:
\begin{equation}
\mathcal{F}^{\mathrm{in}}_1 = \mathcal{F}^{\mathrm{rgb}} + \mathcal{F}^{\mathrm{fused}},
\end{equation}
where $\mathcal{F}^{\mathrm{fused}}$ is the MoFE output and $\mathcal{F}^{\mathrm{in}}_1$ is the input to the first transformer block. In implementation, the weighted joint high-frequency expert output initializes the frequency prior $\mathcal{P}_0$ for subsequent RCFAs:
\begin{equation}
\mathcal{P}_0 = \alpha_4 \widehat{\mathcal{C}}_4,
\end{equation}
where $\mathcal{P}_0$ denotes the initial candidate frequency prior and $\widehat{\mathcal{C}}_4$ is the enhanced joint high-frequency cross-modal frequency feature.

The expert-driven frequency-aware design enables the model to dynamically emphasize relevant subband interactions across modalities and provides information-rich frequency cues. Because its routing weights aggregate image-level frequency evidence rather than explicitly evaluating local cross-modal consistency, MoFE does not itself guarantee that every high-frequency response is reliable. This reliability assessment is deferred to RCFA during stage-wise propagation.

\subsection{Reliability-Calibrated Frequency Adapter}

To mitigate the degradation of fine structures caused by token mixing and downsampling in transformer layers, RCFA focuses on the second step of our frequency-aware design: evaluating and selectively propagating MoFE's candidate high-frequency residuals across transformer stages. Unlike conventional adapters that operate solely on local features, RCFA introduces cross-layer interaction, frequency-aware enhancement, and reliability calibration in a unified and lightweight design, as illustrated in Fig.~\ref{fig:framework}.

At each transformer stage $i$, RCFA receives the current encoder feature $\mathcal{F}_i$ and the incoming frequency prior $\mathcal{P}_{i-1}$, aligning the latter to the spatial size and channel dimension of $\mathcal{F}_i$ when necessary. Here $i$ indexes the RCFA insertion stage and $\mathcal{F}_i \in \mathbb{R}^{C_i \times H_i \times W_i}$. At the first RCFA, the incoming prior is the initial candidate frequency prior $\mathcal{P}_0$. Thereafter, it is the calibrated frequency prior from the preceding RCFA. After alignment, it is still denoted by $\mathcal{P}_{i-1}$ for simplicity. RCFA then constructs a base residual from the encoder feature and a frequency residual from the incoming frequency prior.
The high-frequency responses are obtained by average-pooling subtraction:
\begin{align}
\mathcal{R}^{\mathrm{hf}}_i &=
\mathcal{F}_i - \mathrm{AvgPool}_{3 \times 3}(\mathcal{F}_i),\\
\mathcal{D}^{\mathrm{hf}}_i &=
\mathcal{P}_{i-1} - \mathrm{AvgPool}_{3 \times 3}(\mathcal{P}_{i-1}),
\end{align}
where $\mathcal{R}^{\mathrm{hf}}_i$ and $\mathcal{D}^{\mathrm{hf}}_i$ denote the high-frequency responses of the encoder feature and incoming frequency prior, respectively, and $\mathrm{AvgPool}_{3 \times 3}(\cdot)$ serves as a low-pass filter to isolate high-frequency details such as edges and textures.
Both residual branches are implemented by lightweight bottleneck adapters composed of two $1 \times 1$ convolutional layers:
\begin{equation}
\mathcal{A}_{\mathrm{hf}}(\mathcal{X}) =
W_2 \left( \mathrm{GELU} \left( W_1 \left(\mathcal{X} + \mathcal{X}^{\mathrm{hf}}\right) \right) \right),
\end{equation}
where $\mathcal{A}_{\mathrm{hf}}(\cdot)$ denotes the high-frequency residual adapter, $\mathcal{X}$ is a generic input feature, $\mathcal{X}^{\mathrm{hf}}=\mathcal{X}-\mathrm{AvgPool}_{3 \times 3}(\mathcal{X})$ is its high-frequency residual, $W_1 \in \mathbb{R}^{C/r \times C}$ and $W_2 \in \mathbb{R}^{C \times C/r}$ are learnable $1 \times 1$ convolutions, $C$ is the channel number of $\mathcal{X}$, and $r$ is the channel reduction ratio. We thus obtain the base residual $\mathcal{B}^{\mathrm{res}}_i=\mathcal{A}_{\mathrm{hf}}(\mathcal{F}_i)$ and the frequency residual $\mathcal{Q}^{\mathrm{res}}_i=\mathcal{A}_{\mathrm{hf}}(\mathcal{P}_{i-1})$, generated from the encoder feature and incoming frequency prior, respectively.
The initial candidate frequency prior can contain unreliable responses when auxiliary cues are noisy or spatially misaligned with RGB boundaries. RCFA therefore adopts a dual-gate calibration mechanism. The context gate estimates the overall injection strength from the current feature and incoming frequency prior, while the Reliability-Calibrated Gate (RCG) evaluates their high-frequency consistency to suppress unreliable residuals. Given the encoder high-frequency response $\mathcal{R}^{\mathrm{hf}}_i$ and prior response $\mathcal{D}^{\mathrm{hf}}_i$, RCG first computes three bounded reliability cues:
\begin{align}
\mathcal{E}^{\mathrm{enc}}_i &= \operatorname{Mean}_c(|\mathcal{R}^{\mathrm{hf}}_i|),\\
\mathcal{E}^{\mathrm{prior}}_i &= \operatorname{Mean}_c(|\mathcal{D}^{\mathrm{hf}}_i|),
\end{align}
where $\mathcal{E}^{\mathrm{enc}}_i$ and $\mathcal{E}^{\mathrm{prior}}_i$ are channel-averaged high-frequency energy maps for the encoder and prior responses, respectively, $|\cdot|$ denotes the element-wise absolute value, and $\operatorname{Mean}_c(\cdot)$ denotes channel-wise averaging. Based on these energies, RCG derives consistency, balance, and difference cues:
\begin{align}
\rho^{\mathrm{con}}_i &= \frac{1}{2}\left(1 + \operatorname{CosSim}_c(\mathcal{R}^{\mathrm{hf}}_i,\mathcal{D}^{\mathrm{hf}}_i)\right),\\
\rho^{\mathrm{bal}}_i &= 1 - \frac{\left|\mathcal{E}^{\mathrm{enc}}_i-\mathcal{E}^{\mathrm{prior}}_i\right|}{\mathcal{E}^{\mathrm{enc}}_i+\mathcal{E}^{\mathrm{prior}}_i+\epsilon},\\
\rho^{\mathrm{diff}}_i &= \exp\left(-\frac{\operatorname{Mean}_c\left(\left|\mathcal{R}^{\mathrm{hf}}_i-\mathcal{D}^{\mathrm{hf}}_i\right|\right)}{\mathcal{E}^{\mathrm{enc}}_i+\mathcal{E}^{\mathrm{prior}}_i+\epsilon}\right),
\end{align}
where $\rho^{\mathrm{con}}_i$, $\rho^{\mathrm{bal}}_i$, and $\rho^{\mathrm{diff}}_i$ denote the high-frequency consistency, energy-balance, and difference-suppression cues, respectively, $\operatorname{CosSim}_c(\cdot,\cdot)$ computes cosine similarity along the channel dimension, $\epsilon$ is a small constant for numerical stability, and $\exp(\cdot)$ is the exponential function. These cues are concatenated and transformed by a lightweight convolutional gate:
\begin{equation}
\mathcal{G}^{\mathrm{rel}}_i = 1 + \eta \left(2\sigma\left(\phi([\rho^{\mathrm{con}}_i,\rho^{\mathrm{bal}}_i,\rho^{\mathrm{diff}}_i])\right)-1\right),
\end{equation}
where $\mathcal{G}^{\mathrm{rel}}_i$ is the reliability calibration weight, $\phi(\cdot)$ denotes two lightweight convolutional layers, $[\cdot]$ denotes channel-wise concatenation of the three cue maps, $\sigma(\cdot)$ is the sigmoid function, and $\eta$ is a learnable residual scale initialized to zero. The other part of the dual-gate mechanism is a lightweight context gate $\psi(\mathcal{F}_i+\mathcal{P}_{i-1})$, where $\psi(\cdot)$ is implemented by global average pooling followed by two $1 \times 1$ convolutions and a sigmoid activation. The final adapter residual is formulated as:
\begin{equation}
\mathcal{F}^{\mathrm{res}}_i =
\mathcal{B}^{\mathrm{res}}_i +
\gamma_i \, \psi(\mathcal{F}_i+\mathcal{P}_{i-1}) \odot
\mathcal{G}^{\mathrm{rel}}_i \odot \mathcal{Q}^{\mathrm{res}}_i,
\end{equation}
where $\mathcal{F}^{\mathrm{res}}_i$ is the residual injected into the $i$-th transformer block, $\gamma_i$ is a learnable frequency scale initialized to zero, $\odot$ denotes element-wise multiplication with broadcasting when necessary, $\psi(\mathcal{F}_i+\mathcal{P}_{i-1})$ is the context gate, $\mathcal{G}^{\mathrm{rel}}_i$ is the Reliability-Calibrated Gate, and $\mathcal{Q}^{\mathrm{res}}_i$ is the frequency residual. The two gates jointly form the dual-gate calibration mechanism, enabling the model to learn both how strongly and how reliably the frequency residual should be injected. With $\gamma_i$ initialized to zero, the frequency-residual branch is initially disabled and is gradually activated during training to enhance useful frequency residuals and suppress unreliable ones.
Finally, the calibrated residual is injected before the original transformer block:
\begin{equation}
\mathcal{F}^{\mathrm{out}}_i = \mathrm{Block}_i\left(\mathcal{F}_i + \mathcal{F}^{\mathrm{res}}_i\right),
\end{equation}
where $\mathrm{Block}_i(\cdot)$ denotes the frozen transformer block of the adopted SAM backbone at stage $i$, and $\mathcal{F}^{\mathrm{out}}_i$ is its output feature. The residual $\mathcal{F}^{\mathrm{res}}_i$ is propagated as the calibrated frequency prior $\mathcal{P}_i$ for the next RCFA.
The process is repeated hierarchically across transformer stages. By leveraging current encoder features and reliability-calibrated frequency priors, RCFA preserves spatial precision and enhances boundary sensitivity.

\subsection{Hypernetwork-guided Semantic-Structural Decoder}

To produce accurate and structure-preserving saliency maps, we use a lightweight Hypernetwork-guided Semantic-Structural Decoder (HSSD) to convert reliability-calibrated encoder features into precise object masks. Rather than introducing another independent fusion objective, HSSD comprises two lightweight decoding paths: a semantic path preserves the adopted SAM backbone's hypernetwork-based mask prior, and a structural path restores dense boundary details. The two embeddings are fused at the feature level inside the mask decoder.

Let $\{\mathcal{F}_{i}^{mix}\}_{i=1}^{4}$ denote the four encoder-stage features collected for decoding. At stages equipped with RCFA, $\mathcal{F}_{i}^{mix}=\mathcal{F}^{\mathrm{out}}_i$. At the remaining stage, $\mathcal{F}_{i}^{mix}$ is the output of the corresponding frozen block of the adopted SAM backbone. A lightweight multi-scale neck first aligns their resolutions and channels, producing $\{\widehat{\mathcal{F}}_i\}_{i=1}^{4}$. We denote the aligned feature levels used by the decoder as $\widehat{\mathcal{F}}_{32}$, $\widehat{\mathcal{F}}_{64}$, and $\widehat{\mathcal{F}}_{128}$ according to their spatial resolutions. These aligned features are then directed into two complementary decoding paths. The semantic path employs an FPN-style aggregation and the mask decoder upscaling path of the adopted SAM backbone to produce mask embeddings:
\begin{align}
\mathcal{F}_{\mathrm{FPN}} &= \mathrm{FPN}(\{\widehat{\mathcal{F}}_i\}),\\
\mathcal{U}_{\mathrm{SAM}} &= \mathrm{Upscale}_{\mathrm{SAM}}(\mathcal{F}_{\mathrm{FPN}}).
\end{align}
where $\mathcal{F}_{\mathrm{FPN}}$ is the feature pyramid output, $\mathrm{FPN}(\cdot)$ denotes the SAM FPN neck, $\mathrm{Upscale}_{\mathrm{SAM}}(\cdot)$ denotes the upscaling path in the SAM mask decoder, and $\mathcal{U}_{\mathrm{SAM}}$ is the semantic mask embedding.

The structural path is implemented with a lightweight Mamba-based decoder to recover dense boundary details that may be weakened by the semantic path. It takes three decoder feature levels at resolutions $32 \times 32$, $64 \times 64$, and $128 \times 128$, and progressively projects and fuses them via top-down refinement:
\begin{align}
\mathcal{O}_{32} &= \omega(\widehat{\mathcal{F}}_{32}),\\
\mathcal{O}_{64} &= \omega\!\bigl(\widehat{\mathcal{F}}_{64} + \mathrm{Up}(\mathcal{O}_{32})\bigr),\\
\mathcal{O}_{128} &= \omega\!\bigl(\widehat{\mathcal{F}}_{128} + \mathrm{Up}(\mathcal{O}_{64})\bigr).
\end{align}
where $\mathcal{O}_{32}$, $\mathcal{O}_{64}$, and $\mathcal{O}_{128}$ are progressively refined structural features at the corresponding resolutions, $\omega(\cdot)$ is a $3 \times 3$ convolution followed by normalization and activation, and $\mathrm{Up}(\cdot)$ denotes bilinear upsampling to the next higher resolution. A final VSS block~\cite{liu2024vmamba} is then applied to model long-range dependencies:
\begin{equation}
\mathcal{U}_{\mathrm{Mamba}} = \mathrm{VSS}(\mathcal{O}_{128}).
\end{equation}
where $\mathrm{VSS}(\cdot)$ denotes the visual state-space block used in the Mamba decoder, and $\mathcal{U}_{\mathrm{Mamba}}$ is the structural embedding.

Instead of directly adding two prediction logits, the mask decoder adaptively fuses the semantic embedding and the structural embedding through a lightweight gate:
\begin{align}
\mathcal{G}^{\mathrm{dec}} &= \sigma\!\left(\varphi([\mathcal{U}_{\mathrm{SAM}},\mathcal{U}_{\mathrm{Mamba}}])\right),\\
\mathcal{U}_{\mathrm{fuse}} &=
\mathcal{G}^{\mathrm{dec}} \odot \mathcal{U}_{\mathrm{SAM}}
+ (1-\mathcal{G}^{\mathrm{dec}}) \odot \mathcal{U}_{\mathrm{Mamba}},
\end{align}
where $\mathcal{G}^{\mathrm{dec}}$ is the decoder fusion gate, $\varphi(\cdot)$ is a lightweight $1 \times 1$ convolutional gate, $\sigma(\cdot)$ is the sigmoid activation, $[\cdot]$ denotes channel-wise concatenation, $\odot$ denotes element-wise multiplication, and $\mathcal{U}_{\mathrm{fuse}}$ is the fused semantic-structural embedding. This adaptive fusion allows the semantic embedding to dominate coherent object regions while retaining structural responses near local transitions, avoiding the fixed trade-off of direct feature addition. The final saliency map is then generated by applying the mask hypernetwork to the fused embedding:
\begin{equation}
\mathcal{S} = \mathrm{Sigmoid}\left(\mathcal{H}_{\mathrm{mask}}(\mathcal{U}_{\mathrm{fuse}})\right).
\end{equation}
where $\mathcal{H}_{\mathrm{mask}}(\cdot)$ denotes the SAM mask hypernetwork and $\mathcal{S}$ is the predicted saliency map. Meanwhile, the mask hypernetwork weights are also applied to the Mamba structural embedding to produce an auxiliary edge prediction for boundary supervision.

\subsection{Loss Function}

To effectively train HSSD, distinct loss functions independently supervise semantic completeness and structural precision. Specifically, the Mamba-based structural path, tasked with capturing fine-grained spatial details and boundary precision, is supervised using Dice loss~\cite{zhao2020rethinking}. The edge target is generated from the boundary of the ground-truth saliency mask, and the Dice loss robustly addresses class imbalance, notably improving accuracy along object boundaries.

For the semantic mask decoder inherited from the adopted SAM backbone, which emphasizes coarse but semantically coherent segmentation, a combined loss function incorporating both Dice and Intersection-over-Union (IoU) losses~\cite{rahman2016optimizing} is employed to simultaneously enforce region consistency and spatial overlap accuracy:
\begin{equation}
\ell_{\mathrm{mask}} = \ell_{\mathrm{dice}} + \ell_{\mathrm{iou}}.
\end{equation}
where $\ell_{\mathrm{mask}}$ is the semantic mask loss, and $\ell_{\mathrm{dice}}$ and $\ell_{\mathrm{iou}}$ denote the Dice and IoU losses computed between the predicted saliency map and the ground-truth mask, respectively.

Consequently, the overall training objective integrates these complementary losses:
\begin{equation}
\mathcal{L} = \ell_{\mathrm{mask}} + \ell_{\mathrm{edge}},
\end{equation}
where $\mathcal{L}$ denotes the total training loss, $\ell_{\mathrm{edge}} = \ell_{\mathrm{dice}}$ explicitly supervises the Mamba-based decoder's edge predictions, while $\ell_{\mathrm{mask}}$ optimizes the semantic decoding path of the adopted SAM backbone.

Through the complementary loss function, each decoder branch is guided to its specialized task, collectively contributing to the generation of accurate, robust, and structurally coherent saliency predictions.

\section{Experiment}
\subsection{Experiment Setup}

\textbf{Datasets and Evaluation Metrics.}
We evaluate our proposed method across three categories of multi-modal salient object detection benchmarks: RGB-D, RGB-T, and RGB-NIR datasets.
For RGB-D SOD, experiments are conducted on four widely used datasets in the final release, including NJU2K~\cite{ju2014depth}, NLPR~\cite{peng2014rgbd}, SSD~\cite{SSD}, and STERE~\cite{niu2012leveraging}. Following recent protocols~\cite{fan2020rethinking,chen2020progressively}, we use 1,485 samples from NJU2K and 700 from NLPR for training. The remaining datasets are used exclusively for testing. For RGB-T SOD, we conduct experiments on three widely used benchmark datasets: VT821~\cite{wang2018rgb}, VT1000~\cite{tu2019rgb}, and VT5000~\cite{tu2022rgbt}.
Following the common setting in these methods~\cite{fang2023adnet,10778650}, we use 2,500 image pairs from VT5000 for training, while the remaining 2,500, along with VT821 and VT1000, are used for testing. In addition, to assess cross-modal generalization, we directly apply the trained RGB-T weights to an RGB-NIR dataset~\cite{song2020deep} for inference, without any additional fine-tuning.

For evaluation, we adopt four widely used metrics: F-measure $(F_{\beta})$, Mean Absolute Error ($\mathcal{M}$), E-measure $(E_m)$, S-measure $(S_m)$.

\noindent\textbf{Implementation Details.} The proposed framework is implemented using the PyTorch platform and trained on a workstation equipped with an Intel Core Ultra 9 285K CPU and NVIDIA RTX 5090 GPU. We initialize the adopted SAM backbone from the officially released checkpoint~\cite{ravi2024sam} and retain it in a frozen state.
All input images are resized to a resolution of $512 \times 512$ to align with the adopted SAM backbone. During training, data augmentation techniques such as random flipping, rotation, and cropping are employed to improve generalization and robustness.
S$^3$AM is trained for 50 epochs with a batch size of 8 using the AdamW optimizer. The initial learning rate is set to $1 \times 10^{-4}$, with a weight decay of $5 \times 10^{-4}$ and betas set to $(0.9, 0.999)$. Gradient clipping is applied with a threshold of 0.5 to stabilize optimization.

\subsection{Quantitative Comparisons}

\begin{figure}[t]
\centering
  \includegraphics[width=\linewidth]{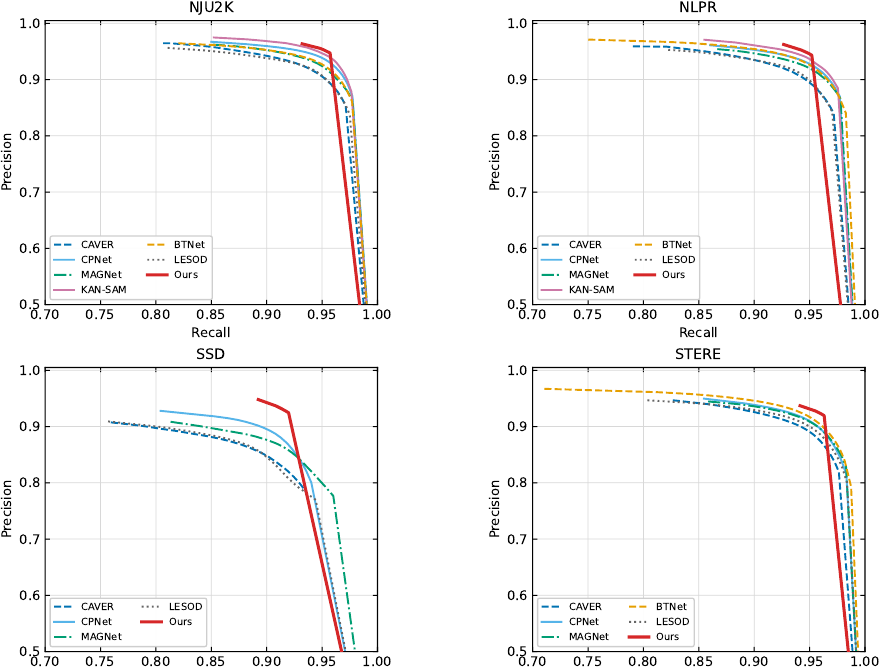}
  \vspace{-6pt}
  \caption{Precision-recall curves of competitive RGB-D salient object detection methods on the NJU2K, NLPR, SSD, and STERE datasets.}
  \label{fig:rgbdpr}
\end{figure}

\noindent\textbf{RGB-D Comparisons.} We compare S$^3$AM against six RGB-D SOD methods, including BTNet~\cite{ren2025bio}, KAN-SAM~\cite{kansam}, CAVER~\cite{caver}, CPNet~\cite{cpnet}, MAGNet~\cite{magnet}, and LESOD~\cite{lesod}. As summarized in Table~\ref{tab:rgbd}, S$^3$AM obtains the highest $F_{\beta}$ and the lowest MAE on all four benchmarks. Compared with the strongest prior $F_{\beta}$ results, the relative improvements are 0.4\%, 1.0\%, 0.3\%, and 0.4\% on NJU2K, NLPR, SSD, and STERE, respectively. The corresponding MAE decreases are 0.003, 0.002, 0.002, and 0.001. Figure~\ref{fig:rgbdpr} further shows favorable precision at high recall, particularly on SSD and STERE. While individual PR curves may cross at particular thresholds, the consistently stronger $F_{\beta}$ and MAE results show that S$^3$AM maintains a more favorable overall precision-recall balance across varied depth quality and scene structures.

\noindent\textbf{RGB-T Comparisons.} We compare S$^3$AM with six RGB-T SOD methods, including CMDBIF~\cite{CMDBIF-Net}, CAVER~\cite{caver}, SMR-Net~\cite{xiao2025smr}, LAFB~\cite{lfab}, UMINet~\cite{uminet}, and LESOD~\cite{lesod}. Table~\ref{tab:rgbt} shows that S$^3$AM achieves the highest $F_{\beta}$ and $S_m$ on all three benchmarks, with the lowest or tied-lowest MAE. Relative to the strongest competing $F_{\beta}$ scores, the gains are 4.5\% on VT5000, 2.4\% on VT1000, and 3.3\% on VT821. Although the competing methods attain comparable $E_m$ on VT5000 and VT1000, the consistently stronger $F_{\beta}$, $S_m$, and MAE results indicate more reliable overall saliency prediction in thermal scenarios.

\noindent\textbf{RGB-NIR Comparisons.} To assess cross-modal generalization, we evaluate S$^3$AM under a zero-shot protocol on the RGB--NIR dataset, without additional fine-tuning, against three representative RGB-NIR methods: RC~\cite{RC}, DCL~\cite{dcl}, and SOD-8S+~\cite{sod8s+}. Table~\ref{tab:nir} reports average F-measure ($F_{\mathrm{avg}}$), maximum F-measure ($F_{\max}$), $F_{\beta}$, MAE ($\mathcal{M}$), $E_m$, and $S_m$. S$^3$AM attains the best result on every metric among the compared methods. In particular, it improves $F_{\beta}$ by 23.6\% and reduces MAE from 0.061 to 0.020 over the strongest compared method, supporting the transferability of the proposed adaptation to NIR inputs.

\begin{table}[t]
\scriptsize
\setlength\tabcolsep{3pt}
\centering
\caption{Comparison with different methods on RGB-D datasets. Best in bold.}
\resizebox{\linewidth}{!}{
\begin{tabular}{lr|ccccccc}
\toprule
& Metric
& \makecell{CAVER\\ \cite{caver}} 
& \makecell{CPNet\\ \cite{cpnet}} 
& \makecell{MAGNet\\ \cite{magnet}}
& \makecell{KAN-SAM\\ \cite{kansam}} 
& \makecell{BTNet\\ \cite{ren2025bio}} 
& \makecell{LESOD\\ \cite{lesod}} 
& \makecell{Ours} \\
\midrule

\multirow{4}{*}{\rotatebox{90}{NJU2K}} 
& $F_{\beta}\uparrow$        & 0.901 & 0.923 & 0.912 & \second{0.935}&0.917 & 0.898 & \best{0.939} \\
& $\mathcal{M}\downarrow$    & 0.031  & \second{0.025} & 0.027 & \second{0.022}&\second{0.025} & 0.032 & \best{0.019} \\
& $E_m\uparrow$              & 0.923  & \second{0.935} & 0.928 & \best{0.942}&0.932 & 0.925 & \second{0.939} \\
& $S_m\uparrow$              & 0.921  & \second{0.935} & 0.929 & \best{0.939}&0.932 & 0.922 & \best{0.939} \\
\midrule

\multirow{4}{*}{\rotatebox{90}{NLPR}} 
& $F_{\beta}\uparrow$        & 0.895  & \second{0.918} & 0.910 & \second{0.925}&0.917 & 0.897 & \best{0.934} \\
& $\mathcal{M}\downarrow$    & 0.020  & \second{0.016} & 0.017 & 0.017&\second{0.016} & 0.020 & \best{0.014} \\
& $E_m\uparrow$              & 0.960  & \second{0.970} & 0.964 & 0.967&0.969 & 0.961 & \best{0.974} \\
& $S_m\uparrow$              & 0.929  & 0.939 & 0.938 & 0.939& \second{0.941} & 0.933 & \best{0.944} \\
\midrule

\multirow{4}{*}{\rotatebox{90}{SSD}} 
& $F_{\beta}\uparrow$        & 0.824      & 0.857 & 0.837 &\second{0.889}& -     & 0.819 & \best{0.892} \\
& $\mathcal{M}\downarrow$    & 0.042      & 0.035 & 0.043 &\second{0.026}& -     & 0.048 & \best{0.024} \\
& $E_m\uparrow$              & 0.911      & 0.917 & 0.915 &\second{0.934}& -     & 0.907 & \best{0.936} \\
& $S_m\uparrow$              & 0.878      & 0.892 & 0.884 &\second{0.910}& -     & 0.879 & \best{0.918} \\
\midrule

\multirow{4}{*}{\rotatebox{90}{STERE}} 
& $F_{\beta}\uparrow$        & 0.881  & 0.895 & 0.893 & \second{0.911}&0.902 & 0.878 & \best{0.915} \\
& $\mathcal{M}\downarrow$    & 0.033  & 0.029 & 0.030 & \second{0.025}&0.028 & 0.038 & \best{0.024} \\
& $E_m\uparrow$              & 0.931  & 0.933 & 0.930 & \second{0.937}&\best{0.940} & \second{0.937} & \second{0.937} \\
& $S_m\uparrow$              & 0.913  & 0.920 & 0.922 & \best{0.930}&0.927 & 0.905 & \best{0.930} \\
\bottomrule
\end{tabular}}
\label{tab:rgbd}
\end{table}

\begin{table}[t]
\scriptsize
\setlength\tabcolsep{3pt}
\centering
\caption{Comparison with different methods on RGB-T datasets. Best in bold.}
\resizebox{\linewidth}{!}{
\begin{tabular}{lr|ccccccc}
\toprule
& Metric
& \makecell{CMDBIF\\ \cite{CMDBIF-Net}} 
& \makecell{CAVER\\ \cite{caver}}
& \makecell{SMR-Net\\ \cite{xiao2025smr}} 
& \makecell{LAFB\\ \cite{lfab}} 
& \makecell{UMINet\\ \cite{uminet}}
& \makecell{LESOD\\ \cite{lesod}} 
& \makecell{Ours} \\
\midrule

\multirow{4}{*}{\rotatebox{90}{VT5000}} 
& $F_{\beta}\uparrow$         & 0.846 & 0.849 & \second{0.859} & 0.841 & 0.820& 0.839 & \best{0.898} \\
& $\mathcal{M}\downarrow$       & 0.032 & 0.028 & \second{0.030} & 0.030 & 0.035 &0.032 & \best{0.020} \\
& $E_m\uparrow$         & 0.933 & 0.935 &\second{0.935} & 0.931 & 0.915& 0.926 & \best{0.950} \\
& $S_m\uparrow$         & 0.886 & \second{0.899} &0.891 & 0.893 & 0.882&0.893 & \best{0.921} \\
\midrule

\multirow{4}{*}{\rotatebox{90}{VT1000}} 
& $F_{\beta}\uparrow$         & 0.909 & \second{0.912} & 0.899 & 0.905 & 0.896& 0.897& \best{0.934} \\
& $\mathcal{M}\downarrow$       & 0.019 & \second{0.016} & 0.020 & 0.018 & 0.021 &0.020& \best{0.013} \\
& $E_m\uparrow$     &    \second{0.952} & \second{0.949} & 0.945 & 0.945 & 0.941&0.937 & \best{0.955} \\
& $S_m\uparrow$        & 0.927 & \second{0.938} & 0.924 & 0.932 & 0.926 &0.929 & \best{0.943} \\
\midrule

\multirow{4}{*}{\rotatebox{90}{VT821}} 
& $F_{\beta}\uparrow$         & 0.837 & \second{0.846} & 0.844 & 0.817 & 0.782&0.827 & \best{0.874} \\
& $\mathcal{M}\downarrow$       & 0.032 & \second{0.026} & 0.030 & 0.034 & 0.054&0.033 & \best{0.025} \\
& $E_m\uparrow$         & 0.923 & \second{0.928} & 0.920 & 0.915 & 0.879&0.917 & \best{0.931} \\
& $S_m\uparrow$         & 0.882 & \second{0.897} & 0.888 & 0.884 & 0.905&\second{0.897} & \best{0.910} \\

\bottomrule
\end{tabular}}
\label{tab:rgbt}
\end{table}

\begin{table}[t]
\scriptsize
  \centering
  \caption{Comparison with representative RGB-NIR methods. Best in bold.}
    \begin{tabular*}{\linewidth}{@{\extracolsep{\fill}}c|cccccc}
    \toprule
    Methods & $F_{\mathrm{avg}}\uparrow$ & $F_{\max}\uparrow$ & $F_{\beta}\uparrow$ & $\mathcal{M}\downarrow$ & $E_m\uparrow$ & $S_m\uparrow$ \\
    \midrule
RC~\cite{RC} & 0.664 & 0.736 & 0.442 & 0.148 & 0.810 & 0.724 \\  
     DCL~\cite{dcl} & 0.779 & 0.838 & 0.660 & 0.076 & 0.881 & 0.796  \\
     SOD8s+~\cite{sod8s+} & \second{0.803} & \second{0.850} & \second{0.745} & \second{0.061} & \second{0.894} & \second{0.828} \\
    \midrule
   Ours & \best{0.928} & \best{0.941} & \best{0.921} & \best{0.020}
 & \best{0.964} & \best{0.930}
 \\
    \bottomrule
    \end{tabular*}%
  \label{tab:nir}%
\end{table}

\subsection{Qualitative Comparisons}
\begin{figure}[t]
\centering
  \includegraphics[width=\linewidth]{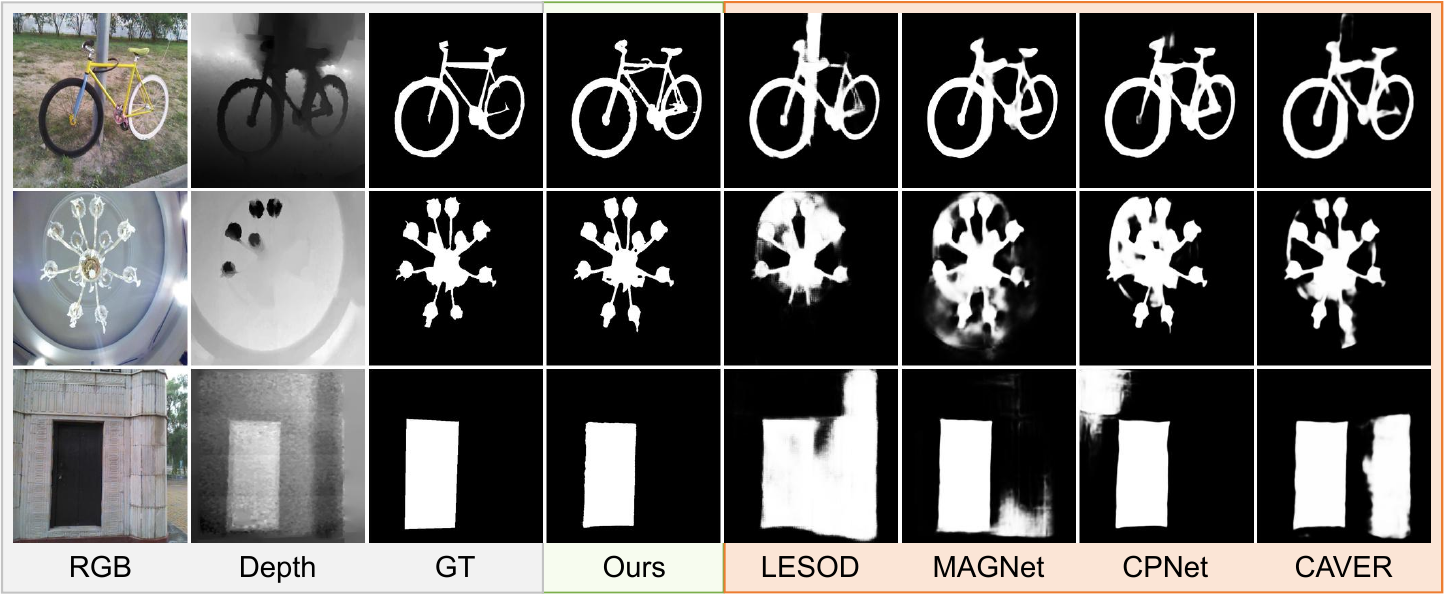}   
  \caption{Qualitative comparison with competitive methods on RGB-D datasets.} 
  \label{fig:rgbdvis}  
\end{figure}

\begin{figure}[!t]
\centering
  \includegraphics[width=\linewidth]{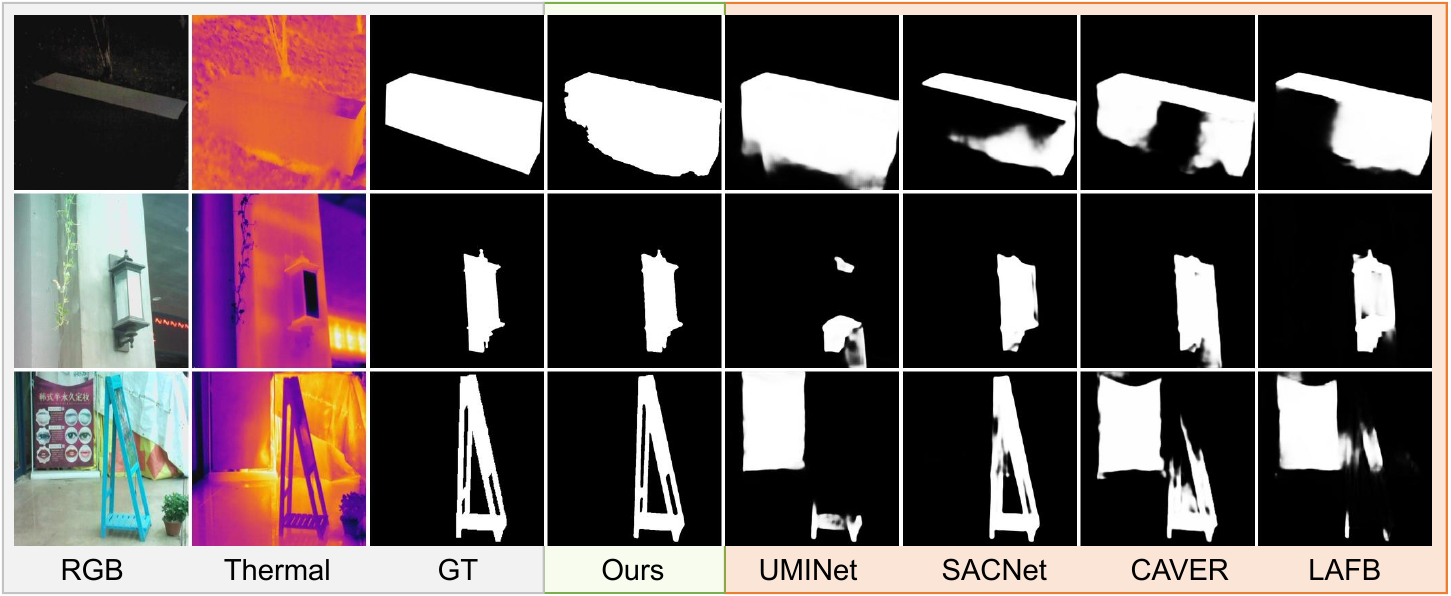}   
  \caption{Qualitative comparison with competitive methods on RGB-T datasets.} 
  \label{fig:rgbtvis}  
\end{figure}

\begin{figure}[!t]
\centering
  \includegraphics[width=\linewidth]{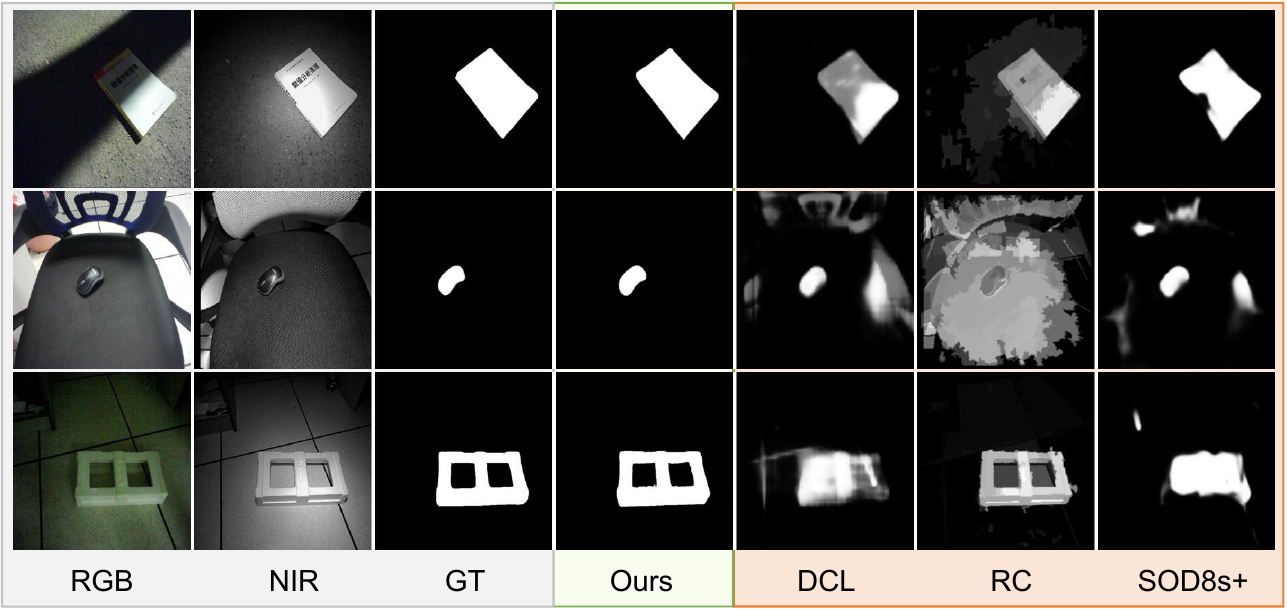}   
  \caption{Qualitative comparison with representative RGB-NIR methods.} 
  \label{fig:rgbnvis}  
\end{figure}

To further assess the effectiveness of the proposed method, we present qualitative comparisons against competitive RGB\mbox{-}D and RGB\mbox{-}T methods (Fig.~\ref{fig:rgbdvis} and Fig.~\ref{fig:rgbtvis}) and representative RGB-NIR methods (Fig.~\ref{fig:rgbnvis}). Across a variety of scenes, S$^3$AM produces sharper boundaries and fewer visual artifacts. In RGB-D examples, it better preserves thin or disconnected target parts while avoiding depth-induced leakage into adjacent regions. In modality-degraded RGB-T scenes, it more accurately delineates targets despite limited or noisy RGB evidence. The RGB-NIR results further show fewer false-positive responses on visually distracting backgrounds. Together, these observations indicate that calibrated frequency cues improve local structural recovery without sacrificing region completeness.

\subsection{Ablation Studies}

\begin{table}
\scriptsize
    \centering
    \setlength{\tabcolsep}{1pt}
    \caption{Component analysis on NJU2K and NLPR. Best in bold.}
    \label{tableabla}
    \begin{tabular}{cccc|cccc|cccc}
        \toprule
        \multirow{2}{*}{Base} & \multirow{2}{*}{MoFE} &\multirow{2}{*}{RCFA}&\multirow{2}{*}{HSSD}& \multicolumn{4}{c|}{NJU2K} & \multicolumn{4}{c}{NLPR}\\
        &&&&$F_{\beta}\uparrow$ & $\mathcal{M}\downarrow$&$E_m\uparrow$ & $S_m\uparrow$&  $F_{\beta}\uparrow$ & $\mathcal{M}\downarrow$&$E_m\uparrow$ & $S_m\uparrow$\\
        \midrule
\ding{51} & &  & & 0.919 & 0.028&  0.928& 0.926&0.901&0.020&0.96&0.926 \\
\ding{51} & \ding{51} & &  & 0.931 & 0.023& 0.929& 0.935&0.917&0.017&0.967&0.936 \\
\ding{51} & & \ding{51} & \ding{51} & 0.931 & \second{0.021}&0.934& \second{0.938}&0.918&0.017&0.965&0.936\\
\ding{51} & \ding{51}& \ding{51} & & \second{0.932} & 0.022&\second{0.937}& 0.937&\second{0.923}&\second{0.016}&\second{0.970}&\second{0.938}\\
\ding{51} & \ding{51}& \ding{51} &\ding{51} & \best{0.939} & \best{0.019}&\best{0.939}& \best{0.939}&\best{0.934}&\best{0.014}&\best{0.974}&\best{0.944}\\
        \bottomrule
    \end{tabular} 
\end{table}

\renewcommand{\best}[1]{#1}
\begin{table}[t]
\scriptsize
    \centering
    \setlength{\tabcolsep}{1.0pt}
    \caption{Dual-gate calibration analysis on NJU2K and NLPR.}
    \label{tab:dualgate}
    \begin{tabular}{l|cccc|cccc}
        \toprule
        \multirow{2}{*}{Variant} & \multicolumn{4}{c|}{NJU2K} & \multicolumn{4}{c}{NLPR} \\
        & $F_{\beta}\uparrow$ & $\mathcal{M}\downarrow$ & $E_m\uparrow$ & $S_m\uparrow$
        & $F_{\beta}\uparrow$ & $\mathcal{M}\downarrow$ & $E_m\uparrow$ & $S_m\uparrow$ \\
        \midrule
        w/o Context Gate & \second{0.938} & \second{0.020} & \second{0.937} & \best{0.939} & \second{0.932} & \best{0.014} & \best{0.974} & \second{0.943} \\
        w/o Reliability Gate & \second{0.938} & \best{0.019} & 0.935 & \best{0.939} & 0.931 & \best{0.014} & \second{0.973} & \second{0.943} \\
        w/o Both Gates & 0.937 & \best{0.019} & 0.931 & \best{0.939} & 0.931 & \best{0.014} & \best{0.974} & \second{0.943} \\
        Full model & \best{0.939} & \best{0.019} & \best{0.939} & \best{0.939}
        & \best{0.934} & \best{0.014} & \best{0.974} & \best{0.944} \\
        \bottomrule
    \end{tabular}
\end{table}

\begin{table}
\scriptsize
    \centering
    \setlength{\tabcolsep}{0.8pt}
\caption{Comparison of fusion strategies on NJU2K and NLPR.}
    \label{tablefusion}
    \begin{tabular}{c|cccc|cccc|cc}
        \toprule
        \multirow{2}{*}{Strategy} & \multicolumn{4}{c|}{NJU2K} & \multicolumn{4}{c|}{NLPR}&\multirow{2}{*}{\makecell{Params.\\(M)}}&\multirow{2}{*}{\makecell{FLOPs\\(G)}}\\
        &$F_{\beta}\uparrow$ & $\mathcal{M}\downarrow$&$E_m\uparrow$ & $S_m\uparrow$&  $F_{\beta}\uparrow$ & $\mathcal{M}\downarrow$&$E_m\uparrow$ & $S_m\uparrow$&&\\
        \midrule
 Early& \second{0.931} & 0.021&0.934& \second{0.938}&0.918&\second{0.017}&0.965&0.936&\best{223.8}&\best{420.1} \\
  Middle& \best{0.937} & \second{0.020}& \second{0.935}& \best{0.941}&\best{0.923}&\best{0.016}&\best{0.969}&\best{0.940}&438.7&816.7 \\
Late & \best{0.937} & \best{0.019}&\best{0.939}& \best{0.941}&\second{0.920}&\best{0.016}&\second{0.966}&\second{0.937}&\second{438.6}&\second{816.7}\\
        \bottomrule
    \end{tabular} 
\end{table}

\begin{figure}[t]
\centering
  \includegraphics[width=\linewidth]{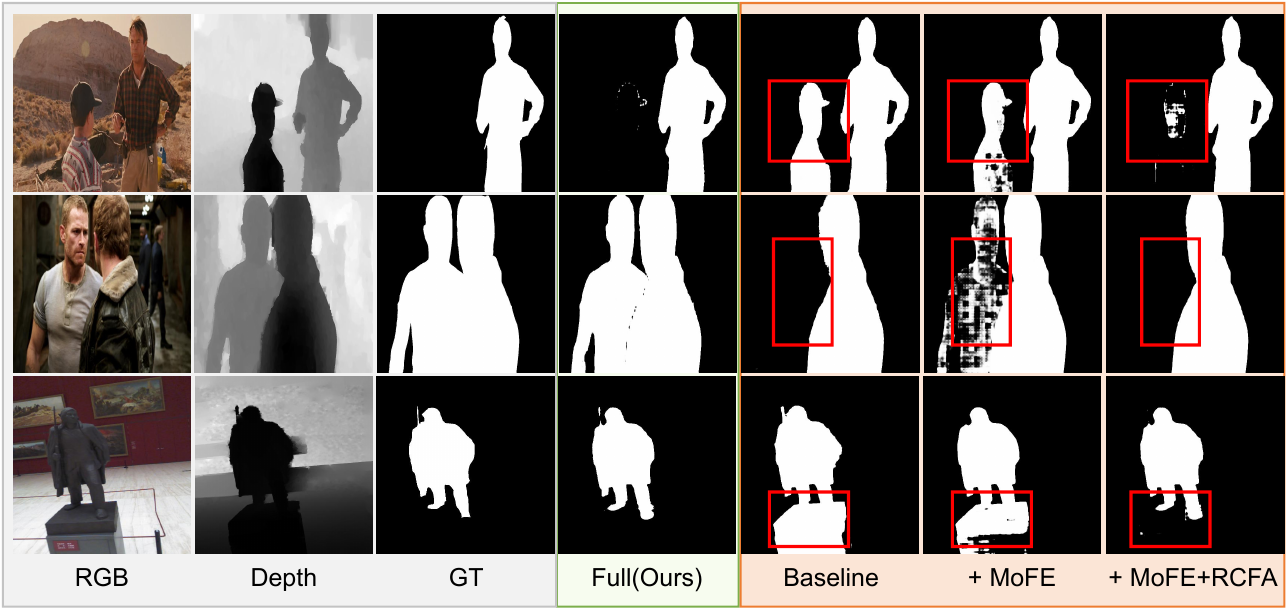}   
  \caption{Qualitative ablation results of the proposed components.} 
  \label{fig:abavis}  
\end{figure}

\begin{figure}[t]
\centering
{\setlength{\fboxsep}{0pt}\setlength{\fboxrule}{0.45pt}
\subfloat[Baseline]{\fcolorbox{gray}{white}{\includegraphics[width=0.22\linewidth]{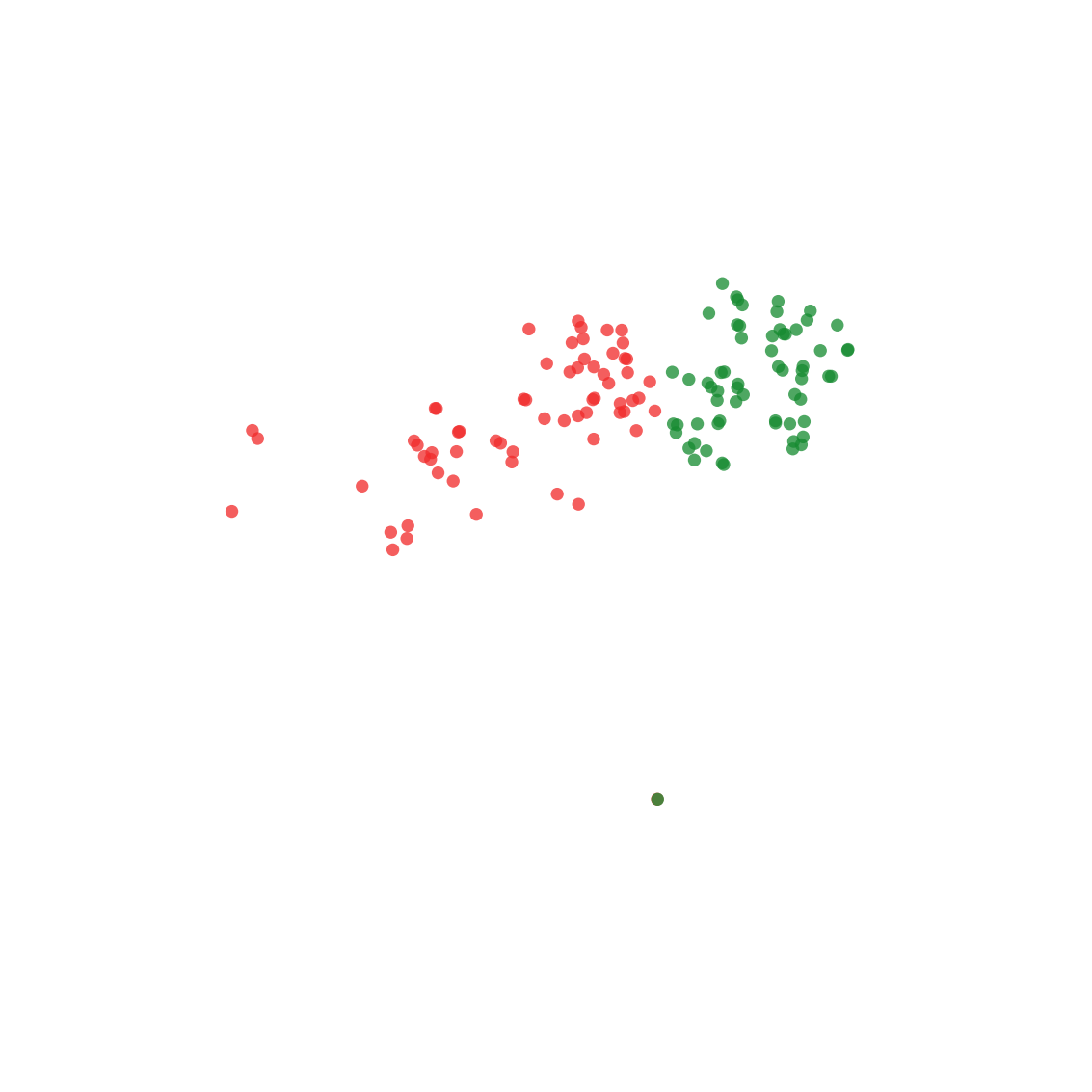}}}\hfill
\subfloat[MoFE]{\fcolorbox{gray}{white}{\includegraphics[width=0.22\linewidth]{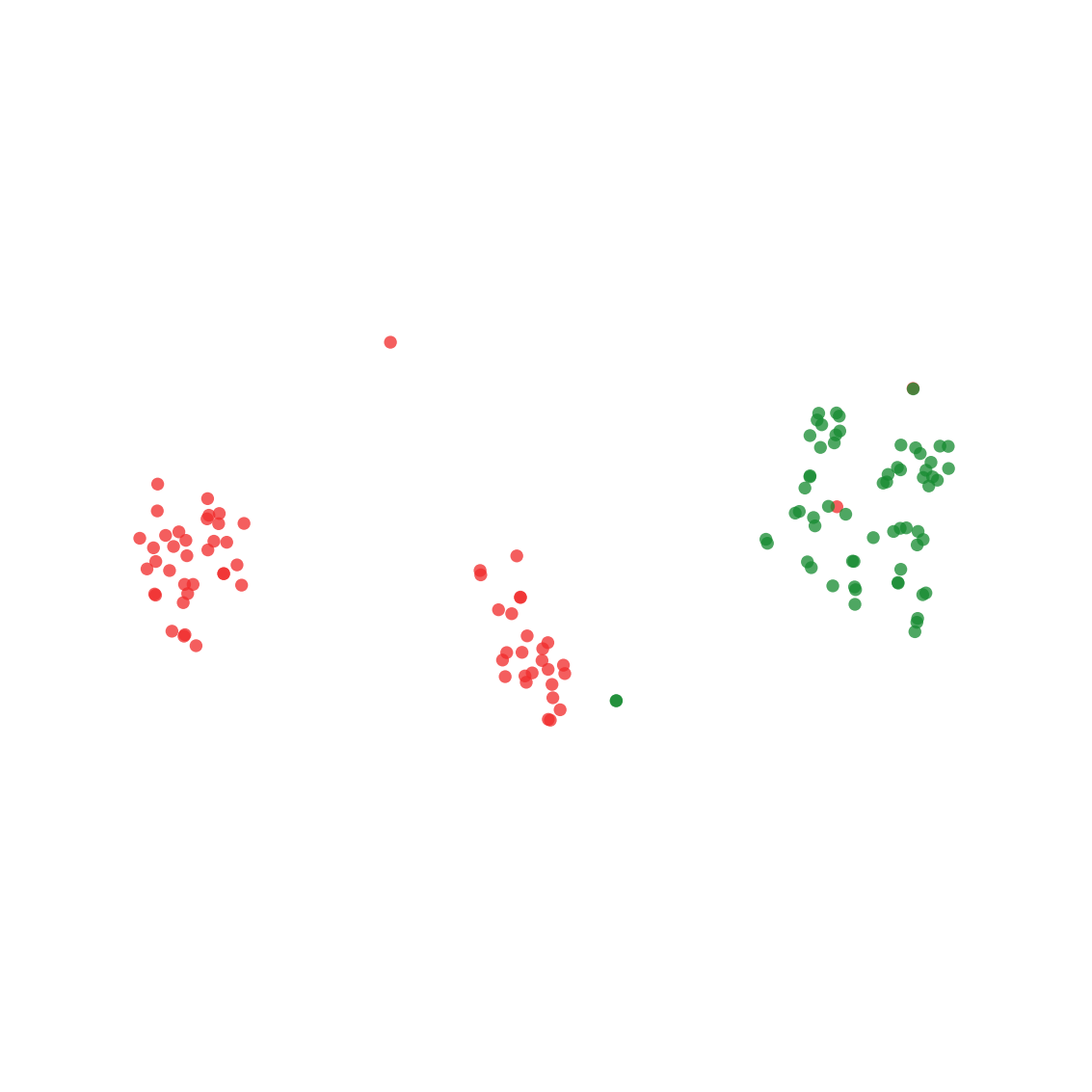}}}\hfill
\subfloat[MoFE+RCFA]{\fcolorbox{gray}{white}{\includegraphics[width=0.22\linewidth]{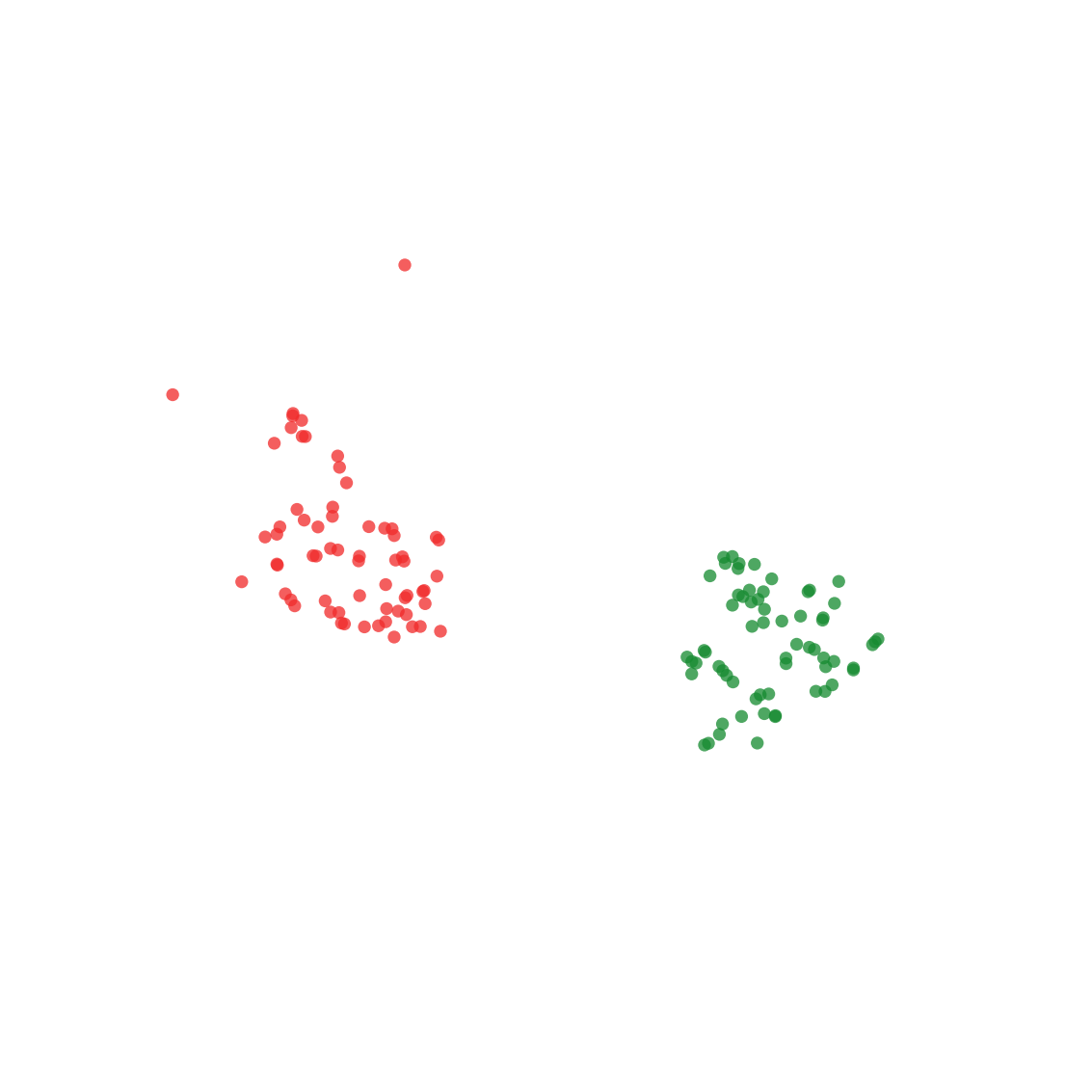}}}\hfill
\subfloat[Full]{\fcolorbox{gray}{white}{\includegraphics[width=0.22\linewidth]{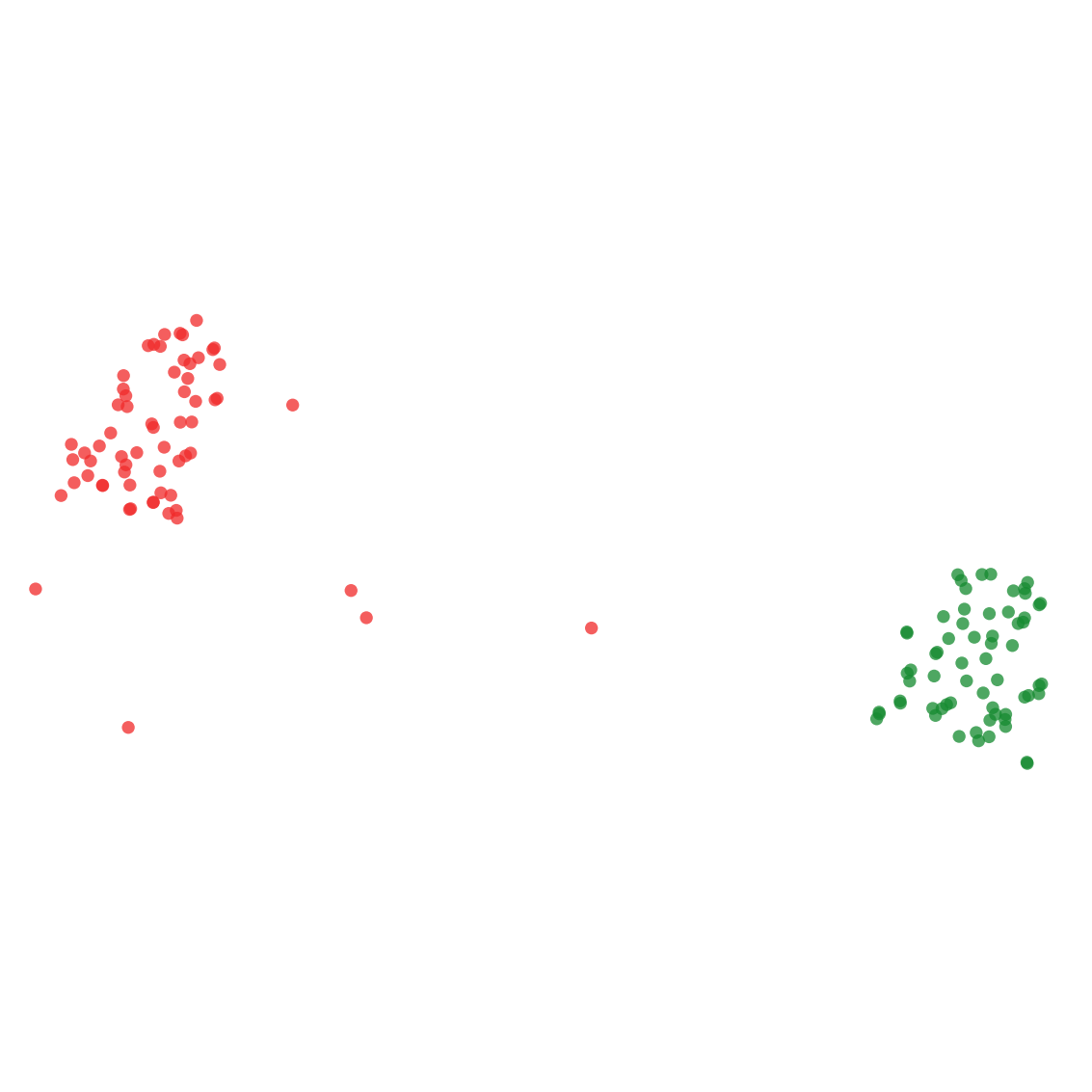}}}}
\caption{t-SNE visualization of token features from a representative RGB-D sample. Red and green markers denote background and foreground token features, respectively. The proposed components progressively improve inter-class separation and intra-class compactness.}
\label{fig:tsne}
\end{figure}

\begin{table}
\scriptsize
    \centering
    \setlength{\tabcolsep}{1.5pt}
\caption{Comparison of model complexity and inference efficiency.}
\label{tab:complex}
    \begin{tabular}{cccc|cccc}
        \toprule
        Base & MoFE & RCFA & HSSD & \makecell{Model\\Params. (M)} & \makecell{Tunable\\Params. (M)} & FLOPs (G) & FPS \\
        \midrule
\ding{51} & &  & & 218.89 & 6.73 & 412.5 & 52.4 \\
\ding{51} & \ding{51} & & & 219.19 & 7.03 & 420.6 & 49.2 \\
\ding{51} & \ding{51}& \ding{51} & & 224.06 & 11.90 & 427.0 & 47.6\\
\ding{51} & \ding{51}& \ding{51} &\ding{51} & 224.36 & 12.20 & 428.1 & 45.8\\
        \bottomrule
    \end{tabular} 
\end{table}

Table~\ref{tableabla} reports the component ablations. The first row is the SAM baseline, where only the decoder and FPN neck are trainable and the encoder is frozen. When RCFA is evaluated without MoFE, we use the additive prior $\mathcal{P}_0=\mathcal{F}^{\mathrm{rgb}}+\mathcal{F}^{\mathrm{aux}}$ to isolate expert-based frequency construction. Introducing MoFE raises $F_{\beta}$ from 0.919 to 0.931 on NJU2K and from 0.901 to 0.917 on NLPR, while reducing MAE by 0.005 and 0.003, respectively. Adding RCFA to MoFE further improves $F_{\beta}$ to 0.932 and 0.923, showing that the initial frequency cues benefit from stage-wise calibration rather than direct propagation alone. The full configuration reaches 0.939 and 0.934, with the lowest MAE on both datasets. The comparison between the MoFE-free RCFA--HSSD configuration and the full model also shows that expert-based frequency construction remains beneficial after the decoder is introduced.

To isolate the dual-gate calibration mechanism within RCFA, we remove the Context Gate and Reliability-Calibrated Gate separately or jointly while retaining all other components. As shown in Table~\ref{tab:dualgate}, removing either gate reduces $F_{\beta}$ from 0.939 to 0.938 on NJU2K and from 0.934 to 0.932 or 0.931 on NLPR. Removing both gates further decreases $F_{\beta}$ to 0.937 and 0.931. The larger degradation on NLPR indicates that contextual injection control and encoder--prior reliability calibration provide complementary cues when auxiliary information is less stable.

Figure~\ref{fig:tsne} visualizes token features from a representative RGB-D sample using t-SNE. The baseline shows dispersed groups and local inter-class mixing. MoFE improves the initial separation, RCFA increases feature compactness, and the full model yields the clearest foreground-background separation.

Figure~\ref{fig:abavis} presents qualitative ablations. From left to right, the fourth through seventh prediction columns show the full model, the baseline, the model with MoFE, and the model with MoFE and RCFA. The progressive refinement sharpens object boundaries while preserving semantic completeness.
\subsection{Complexity Analysis}
Table~\ref{tab:complex} reports total and trainable parameters, FLOPs, and throughput at $512 \times 512$ resolution on an RTX 5090. The full model contains 224.36M total and 12.20M trainable parameters (5.4\%), with 428.1G FLOPs and 45.8 FPS. It also runs at about 5 FPS on an NVIDIA Orin edge computing device. Compared with the baseline, MoFE, RCFA, and HSSD only add 0.30M, 4.87M, and 0.30M trainable parameters, respectively. KAN-SAM~\cite{kansam}, a SAM-based adaptation method, reports 643.588M parameters, 1824G FLOPs, and 4 FPS, corresponding to about 2.9$\times$ parameters and 4.3$\times$ FLOPs of S$^3$AM. HyPSAM (SAMv2)~\cite{hypsam}, a two-stage Swin-B and ViT-H pipeline, reports 817.6M parameters, 3033.4G FLOPs, and 3 FPS. These costs illustrate the efficiency benefit of avoiding an additional prompt-generation and refinement stage.
\subsection{Fusion Strategy Analysis}
Table~\ref{tablefusion} compares early, middle, and late fusion. Early fusion adds $\mathcal{F}^{\mathrm{rgb}}$ and $\mathcal{F}^{\mathrm{aux}}$ before the first transformer block of a shared SAM encoder, without MoFE, thereby isolating the fusion stage. Middle fusion uses two encoder branches with stage-wise interaction, whereas late fusion combines the two representations only at the final stage. Middle fusion improves $F_{\beta}$ from 0.931 to 0.937 on NJU2K and from 0.918 to 0.923 on NLPR, but requires 438.7M parameters and 816.7G FLOPs. Late fusion has comparable overhead, with 438.6M parameters and 816.7G FLOPs, compared with 223.8M parameters and 420.1G FLOPs for early fusion. Thus, early fusion offers a favorable efficiency-performance trade-off, while the gains from the dual-encoder alternatives remain modest relative to their nearly doubled cost.

\section{Conclusion}
In this paper, we proposed S$^3$AM, a single-stream framework for multi-modal salient object detection built upon the adopted SAM backbone. By coupling early cross-modal fusion with reliability-calibrated frequency adaptation, S$^3$AM retains informative structural details while limiting the propagation of unreliable auxiliary cues.
To adaptively model multi-frequency and cross-modal interactions, MoFE constructs initial frequency cues at the early stage. RCFA then evaluates, calibrates, and selectively propagates the calibrated residual throughout the encoder through dual-gate control of injection strength and RGB--auxiliary high-frequency reliability. Finally, HSSD converts the reliability-calibrated features into saliency maps by combining the adopted SAM backbone's hypernetwork-based semantic mask prior with structural detail recovery.
Experiments on RGB-D, RGB-T, and RGB-NIR benchmarks demonstrate that S$^3$AM achieves competitive performance without duplicating foundation-backbone computation.

\bibliographystyle{IEEEbib}
\bibliography{mybib}

@String(ICME = {Int. Conf. Multimedia and Expo})

@String(AAAI = {AAAI})

@String(ICME  =	{ICME})

@inproceedings{zhang2025dimsod,
  title={DiMSOD: A Diffusion-Based Framework for Multi-Modal Salient Object Detection},
  author={Zhang, Shuo and Huang, Jiaming and Tang, Wenbing and Wu, Yan and Hu, Terrence and Xu, Xiaogang and Liu, Jing},
  booktitle={Proc. AAAI Conf. Artif. Intell.},
  year={2025}
}

@article{wang2022mfgnet,
  title={{MFGNet}: Dynamic modality-aware filter generation for {RGB-T} tracking},
  author={Wang, Xiao and Shu, Xiujun and Zhang, Shiliang and Jiang, Bo and Wang, Yaowei and Tian, Yonghong and Wu, Feng},
  journal={IEEE Trans. Multimedia},
  volume={25},
  pages={4335--4348},
  year={2022},
  publisher={IEEE}
}

@article{lu2025image,
  title={Image retrieval using deep saliency edge feature},
  author={Lu, Zhou and Liu, Guang-Hai and Li, Zuo-Yong and Zhang, Bo-Jian},
  journal={Eng. Appl. Artif. Intell.},
  volume={149},
  pages={110416},
  year={2025},
  publisher={Elsevier}
}

@article{gao2025saliency,
  title={Saliency-Aware Foveated Path Tracing for Virtual Reality Rendering},
  author={Gao, Yang and Li, Wencan and Liang, Shiyu and Hao, Aimin and Tan, Xiaohui},
  journal={IEEE Trans. Vis. Comput. Graph.},
  year={2025},
  publisher={IEEE}
}

@inproceedings{chen2023sam,
  title={{SAM-Adapter}: Adapting segment anything in underperformed scenes},
  author={Chen, Tianrun and Zhu, Lanyun and Deng, Chaotao and Cao, Runlong and Wang, Yan and Zhang, Shangzhan and Li, Zejian and Sun, Lingyun and Zang, Ying and Mao, Papa},
  booktitle={Proc. IEEE/CVF Int. Conf. Comput. Vis.},
  year={2023}
}

@article{ravi2024sam,
  title={{SAM 2}: Segment anything in images and videos},
  author={Ravi, Nikhila and Gabeur, Valentin and Hu, Yuan-Ting and Hu, Ronghang and Ryali, Chaitanya and Ma, Tengyu and Khedr, Haitham and R{\"a}dle, Roman and Rolland, Chloe and Gustafson, Laura and others},
  journal={arXiv preprint arXiv:2408.00714},
  year={2024}
}

@article{zhai2025weakly,
  title={Weakly Supervised {RGBT} salient object detection via {SAM}-Guided Label Optimization and Progressive Cross-modal Cross-scale Fusion},
  author={Zhai, Sulan and Liu, Chengzhuang and Tu, Zhengzheng and Li, Chenglong and Gao, Liuxuanqi},
  journal={Inf. Fusion},
  volume={120},
  pages={103048},
  year={2025},
  publisher={Elsevier}
}

@inproceedings{zhao2020rethinking,
  title={Rethinking dice loss for medical image segmentation},
  author={Zhao, Rongjian and Qian, Buyue and Zhang, Xianli and Li, Yang and Wei, Rong and Liu, Yang and Pan, Yinggang},
  booktitle={Proc. IEEE Int. Conf. Data Mining (ICDM)},
  year={2020},

}

@ARTICLE{10909610,
  author={Liu, Zhengyi and Deng, Sheng and Wang, Xinrui and Wang, Linbo and Fang, Xianyong and Tang, Bin},
  journal={IEEE Trans. Multimedia}, 
  title={SSFam: Scribble Supervised Salient Object Detection Family}, 
  year={2025},
  volume={27},
  number={},
  pages={1988-2000},
  doi={10.1109/TMM.2025.3543092}}

@inproceedings{rahman2016optimizing,
  title={Optimizing intersection-over-union in deep neural networks for image segmentation},
  author={Rahman, Md Atiqur and Wang, Yang},
  booktitle={Proc. Int. Symp. Vis. Comput.},
  year={2016},
  
}

@inproceedings{wu2025every,
  title={Every {SAM} Drop Counts: Embracing Semantic Priors for Multi-Modality Image Fusion and Beyond},
  author={Wu, Guanyao and Liu, Haoyu and Fu, Hongming and Peng, Yichuan and Liu, Jinyuan and Fan, Xin and Liu, Risheng},
  booktitle={Proc. IEEE/CVF Conf. Comput. Vis. Pattern Recognit.},
  year={2025}
}

@article{demirel2010image,
  title={Image resolution enhancement by using discrete and stationary wavelet decomposition},
  author={Demirel, Hasan and Anbarjafari, Gholamreza},
  journal={IEEE Trans. Image Process.},
  volume={20},
  number={5},
  pages={1458--1460},
  year={2010},
  publisher={IEEE}
}

@inproceedings{ju2014depth,
  title={Depth saliency based on anisotropic center-surround difference},
  author={Ju, Ran and Ge, Ling and Geng, Wenjing and Ren, Tongwei and Wu, Gangshan},
  booktitle={Proc. IEEE Int. Conf. Image Process.},
  year={2014}

}

@inproceedings{peng2014rgbd,
  title={{RGBD} salient object detection: A benchmark and algorithms},
  author={Peng, Houwen and Li, Bing and Xiong, Weihua and Hu, Weiming and Ji, Rongrong},
  booktitle={Proc. Eur. Conf. Comput. Vis.},
  year={2014},

}

@inproceedings{niu2012leveraging,
  title={Leveraging stereopsis for saliency analysis},
  author={Niu, Yuzhen and Geng, Yujie and Li, Xueqing and Liu, Feng},
  booktitle={Proc. IEEE Conf. Comput. Vis. Pattern Recognit.},
  year={2012},

}

@article{fan2020rethinking,
  title={Rethinking {RGB-D} salient object detection: Models, data sets, and large-scale benchmarks},
  author={Fan, Deng-Ping and Lin, Zheng and Zhang, Zhao and Zhu, Menglong and Cheng, Ming-Ming},
  journal={IEEE Trans. Neural Netw. Learn. Syst.},
  volume={32},
  number={5},
  pages={2075--2089},
  year={2020},
  publisher={IEEE}
}

@inproceedings{chen2020progressively,
  title={Progressively guided alternate refinement network for {RGB-D} salient object detection},
  author={Chen, Shuhan and Fu, Yun},
  booktitle={Proc. Eur. Conf. Comput. Vis.},
  year={2020},
 
}

@inproceedings{wang2018rgb,
  title={{RGB-T} saliency detection benchmark: Dataset, baselines, analysis and a novel approach},
  author={Wang, Guizhao and Li, Chenglong and Ma, Yunpeng and Zheng, Aihua and Tang, Jin and Luo, Bin},
  booktitle={Proc. Chinese Conf. Image Graph. Technol.},
  year={2018},
 
}

@article{tu2019rgb,
  title={{RGB-T} image saliency detection via collaborative graph learning},
  author={Tu, Zhengzheng and Xia, Tian and Li, Chenglong and Wang, Xiaoxiao and Ma, Yan and Tang, Jin},
  journal={IEEE Trans. Multimedia},
  volume={22},
  number={1},
  pages={160--173},
  year={2019},
  publisher={IEEE}
}

@article{tu2022rgbt,
  title={{RGBT} salient object detection: A large-scale dataset and benchmark},
  author={Tu, Zhengzheng and Ma, Yan and Li, Zhun and Li, Chenglong and Xu, Jieming and Liu, Yongtao},
  journal={IEEE Trans. Multimedia},
  volume={25},
  pages={4163--4176},
  year={2022},
  publisher={IEEE}
}

@article{song2020deep,
  title={Deep domain adaptation based multi-spectral salient object detection},
  author={Song, Shaoyue and Miao, Zhenjiang and Yu, Hongkai and Fang, Jianwu and Zheng, Kang and Ma, Cong and Wang, Song},
  journal={IEEE Trans. Multimedia},
  volume={24},
  pages={128--140},
  year={2020},
  publisher={IEEE}
}

@ARTICLE{10778650,
  author={Tang, Hao and Li, Zechao and Zhang, Dong and He, Shengfeng and Tang, Jinhui},
  journal={IEEE Trans. Pattern Anal. Mach. Intell.}, 
  title={Divide-and-Conquer: Confluent Triple-Flow Network for {RGB-T} Salient Object Detection}, 
  year={2025},
  volume={47},
  number={3},
  pages={1958-1974},
  doi={10.1109/TPAMI.2024.3511621}}

@inproceedings{fang2023adnet,
  title={{ADNet}: An Asymmetric Dual-Stream Network for {RGB-T} Salient Object Detection},
  author={Fang, Yaqun and Hou, Ruichao and Bei, Jia and Ren, Tongwei and Wu, Gangshan},
  booktitle={Proc. ACM Int. Conf. Multimedia Asia},
  year={2023}
}

@article{ma2024segment,
  title={Segment anything in medical images},
  author={Ma, Jun and He, Yuting and Li, Feifei and Han, Lin and You, Chenyu and Wang, Bo},
  journal={Nat. Commun.},
  volume={15},
  number={1},
  pages={654},
  year={2024},
  publisher={Nature Publishing Group UK London}
}

@article{wang2023samrs,
  title={{SAMRS}: Scaling-up remote sensing segmentation dataset with segment anything model},
  author={Wang, Di and Zhang, Jing and Du, Bo and Xu, Minqiang and Liu, Lin and Tao, Dacheng and Zhang, Liangpei},
  journal={Adv. Neural Inf. Process. Syst.},
  volume={36},
  pages={8815--8827},
  year={2023}
}

@article{huo2022real,
  title={Real-time one-stream semantic-guided refinement network for {RGB-thermal} salient object detection},
  author={Huo, Fushuo and Zhu, Xuegui and Zhang, Qian and Liu, Ziming and Yu, Wenchao},
  journal={IEEE Trans. Instrum. Meas.},
  volume={71},
  pages={1--12},
  year={2022},
  publisher={IEEE}
}

@article{xie2023cross,
  title={Cross-modality double bidirectional interaction and fusion network for {RGB-T} salient object detection},
  author={Xie, Zhengxuan and Shao, Feng and Chen, Gang and Chen, Hangwei and Jiang, Qiuping and Meng, Xiangchao and Ho, Yo-Sung},
  journal={IEEE Trans. Circuits Syst. Video Technol.},
  volume={33},
  number={8},
  pages={4149--4163},
  year={2023},
  publisher={IEEE}
}

@article{ren2025bio,
  title={Bio-inspired two-stage network for efficient {RGB-D} salient object detection},
  author={Ren, Peng and Bai, Tian and Sun, Fuming},
  journal={Neural Netw.},
  volume={185},
  pages={107244},
  year={2025},
  publisher={Elsevier}
}

@inproceedings{SSD,
  author    = {Chunbiao Zhu and Ge Li},
  title     = {A Three-Pathway Psychobiological Framework of Salient Object Detection Using Stereoscopic Technology},
  booktitle = {Proc. IEEE Int. Conf. Comput. Vis. Workshops},
  year      = {2017}
}

@article{caver,
  title={{CAVER}: Cross-modal view-mixed transformer for bi-modal salient object detection},
  author={Pang, Youwei and Zhao, Xiaoqi and Zhang, Lihe and Lu, Huchuan},
  journal={IEEE Trans. Image Process.},
  volume={32},
  pages={892--904},
  year={2023},
  publisher={IEEE}
}

@article{cpnet,
  title={Cross-modal fusion and progressive decoding network for {RGB-D} salient object detection},
  author={Hu, Xihang and Sun, Fuming and Sun, Jing and Wang, Fasheng and Li, Haojie},
  journal={Int. J. Comput. Vis.},
  volume={132},
  number={8},
  pages={3067--3085},
  year={2024},
  publisher={Springer}
}

@article{magnet,
  title={{MAGNet}: Multi-scale Awareness and Global fusion Network for {RGB-D} salient object detection},
  author={Zhong, Mingyu and Sun, Jing and Ren, Peng and Wang, Fasheng and Sun, Fuming},
  journal={Knowl.-Based Syst.},
  volume={299},
  pages={112126},
  year={2024},
  publisher={Elsevier}
}

@article{lesod,
  title={{LESOD}: Lightweight and Efficient Network for {RGB-D} Salient Object Detection},
  author={Zhong, Mingyu and Sun, Jing and Wang, Fasheng and Sun, Fuming},
  journal={Pattern Recognit.},
  pages={112103},
  year={2025},
  publisher={Elsevier}
}

@article{sod8s+,
  author  = {Song, S. and Miao, Z. and Yu, H. and Fang, J. and Zheng, K. and Ma, C. and Wang, S.},
  title   = {Deep Domain Adaptation Based Multi-Spectral Salient Object Detection},
  journal = {IEEE Trans. Multimedia},
  volume  = {24},
  pages   = {128--140},
  year    = {2022},
  doi     = {10.1109/TMM.2020.3046868}
}

@article{RC,
  author  = {Cheng, M.-M. and Mitra, N.~J. and Huang, X. and Torr, P.~H.~S. and Hu, S.-M.},
  title   = {Global Contrast Based Salient Region Detection},
  journal = {IEEE Trans. Pattern Anal. Mach. Intell.},
  volume  = {37},
  number  = {3},
  pages   = {569--582},
  year    = {2014}
}

@inproceedings{dcl,
  author    = {Li, G. and Yu, Y.},
  title     = {Deep Contrast Learning for Salient Object Detection},
  booktitle = {Proc. IEEE Conf. Comput. Vis. Pattern Recognit.},
  year      = {2016}
}

@article{uminet,
  author = {Gao, L. and Fu, P. and Xu, M. and Wang, T. and Liu, B.},
  title = {{UMINet}: A unified multi-modality interaction network for {RGB-D} and {RGB-T} salient object detection},
  journal = {Vis. Comput.},
  volume = {40},
  number = {3},
  pages = {1565--1582},
  year = {2024}
}

@article{lfab,
  author = {Wang, K. and Tu, Z. and Li, C. and Zhang, C. and Luo, B.},
  title = {Learning adaptive fusion bank for multi-modal salient object detection},
  journal = {IEEE Trans. Circuits Syst. Video Technol.},
  volume = {34},
  pages = {7344--7358},
  year = {2024}
}

@inproceedings{xiao2025smr,
  title = {{SMR-Net}: Semantic-guided mutually reinforcing network for cross-modal image fusion and salient object detection},
  author = {Xiao, Guobao and Liu, Xinyu and Lin, Zebin and Ming, Rui},
  booktitle = {Proc. AAAI Conf. Artif. Intell.},
  year = {2025}
}

@article{SACNET,
  author = {Wang, K. and Lin, D. and Li, C. and Tu, Z. and Luo, B.},
  title = {Alignment-free {RGBT} salient object detection: Semantics-guided asymmetric correlation network and a unified benchmark},
  journal = {IEEE Trans. Multimedia},
  volume = {26},
  pages = {10692--10707},
  year = {2024}
}

@article{wang2026multi,
  author = {Wang, Qishun and Tu, Zhengzheng and Jiang, Bo and Li, Chenglong and Zhang, Pingping and Tang, Jin},
  title = {Multi-graph cross diffusion attention network for alignment-free {RGBT} video object detection},
  journal = {Int. J. Comput. Vis.},
  volume = {134},
  number = {8},
  pages = {375},
  year = {2026}
}

@article{CMDBIF-Net,
  author = {Xie, Z. and Shao, F. and Chen, G. and Chen, H. and Jiang, Q. and Meng, X. and Ho, Y.-S.},
  title = {Cross-modality double bidirectional interaction and fusion network for {RGB-T} salient object detection},
  journal = {IEEE Trans. Circuits Syst. Video Technol.},
  volume = {33},
  number = {8},
  pages = {4149--4163},
  year = {2023}
}

@inproceedings{kansam,
  author={Li, Xingyuan and Hou, Ruichao and Ren, Tongwei and Wu, Gangshan},
  title={{KAN-SAM}: Kolmogorov-Arnold Network Guided Segment Anything Model for {RGB-T} Salient Object Detection},
  booktitle={Proc. IEEE Int. Conf. Multimedia Expo (ICME)},
  year={2025}
}

@article{hypsam,
  author={Hou, Ruichao and Li, Xingyuan and Ren, Tongwei and Zhou, Dongming and Wu, Gangshan and Cao, Jinde},
  journal={IEEE Trans. Circuits Syst. Video Technol.},
  title={{HyPSAM}: Hybrid Prompt-Driven Segment Anything Model for {RGB-Thermal} Salient Object Detection},
  volume={36},
  number={3},
  pages={2697--2712},
  year={2026},
  doi={10.1109/TCSVT.2025.3613770}
}

@article{liu2026samsod,
  title={{SAMSOD}: Rethinking {SAM} Optimization for {RGB-T} Salient Object Detection},
  author={Liu, Zhengyi and Wang, Xinrui and Fang, Xianyong and Tu, Zhengzheng and Wang, Linbo},
  journal={IEEE Trans. Multimedia},
  year={2026},
  publisher={IEEE}
}

@inproceedings{liu2024vmamba,
  author={Liu, Yue and Tian, Yunjie and Zhao, Yuzhong and Yu, Hongtian and Xie, Lingxi and Wang, Yaowei and Ye, Qixiang and Jiao, Jianbin and Liu, Yunfan},
  title={{VMamba}: Visual State Space Model},
  booktitle={Proc. Adv. Neural Inf. Process. Syst.},
  year={2024}
}

\end{document}